\documentclass[lettersize,journal]{IEEEtran}
\usepackage{booktabs}
\usepackage{multirow}
\usepackage{amsmath,amsfonts}
\usepackage{algorithmic}
\usepackage{algorithm}
\usepackage{array}
\usepackage[caption=false,font=normalsize,labelfont=sf,textfont=sf]{subfig}
\usepackage{textcomp}
\usepackage{stfloats}
\usepackage{url}
\usepackage{verbatim}
\usepackage{graphicx}
\usepackage{cite}
\begin{document}

\title{Learning Orientation Field for OSM-Guided Autonomous Navigation}

\author{Yuming Huang$^{1}$, Wei Gao$^{1}$, Zhiyuan Zhang$^{2}$, \\Maani Ghaffari$^{3}$, Dezhen Song$^{4}$, Cheng-Zhong Xu$^{1}$, and Hui Kong$^{1}$
\thanks{
$^{1}$ Yuming Huang, Wei Gao, Cheng-Zhong Xu and Hui Kong are with the State Key Laboratory of Internet of Things for Smart City (SKL-IOTSC), Faculty of Science and Technology, University of Macau, Macau, China. Email: {yc17907, yc37927\}@um.edu.mo, \{czxu, huikong\}@um.edu.mo}\\
$^{2}$ Zhiyuan Zhang is with the School of Computing and Information Systems, Singapore Management University, Singapore. Email: zhiyuanzhang@smu.edu.sg\\
$^{3}$ Maani Ghaffari is with the Department of Naval Architecture and Marine Engineering and Department of Robotics, University of Michigan, Ann Arbor, MI, USA. Email: maanigj@umich.edu\\ 
$^{4}$ Dezhen Song is with the Department of Robotics, 
Mohamed bin Zayed University of Artificial Intelligence (MBZUAI), UAE. Email: dezhen.song@mbzuai.ac.ae
}}
% \author{Anonymous Authors}



\maketitle

\begin{abstract}
OpenStreetMap (OSM) has gained popularity recently in autonomous navigation due to its public accessibility, lower maintenance costs, and broader geographical coverage. However, existing methods often struggle with noisy OSM data and incomplete sensor observations, leading to inaccuracies in trajectory planning. These challenges are particularly evident in complex driving scenarios, such as at intersections or facing occlusions. To address these challenges, we propose a robust and explainable two-stage framework to learn an Orientation Field (OrField) for robot navigation by integrating LiDAR scans and OSM routes. In the first stage, we introduce the novel representation, OrField, which can provide orientations for each grid on the map, reasoning jointly from noisy LiDAR scans and OSM routes. To generate a robust OrField, we train a deep neural network by encoding a versatile initial OrField and output an optimized OrField. Based on OrField, we propose two trajectory planners for OSM-guided robot navigation, called Field-RRT* and Field-Bezier, respectively, in the second stage by improving the Rapidly Exploring Random Tree (RRT) algorithm and Bezier curve to estimate the trajectories. Thanks to the robustness of OrField which captures both global and local information, Field-RRT* and Field-Bezier can generate accurate and reliable trajectories even in challenging conditions. We validate our approach through experiments on the SemanticKITTI dataset and our own campus dataset. The results demonstrate the effectiveness of our method, achieving superior performance in complex and noisy conditions. Our code for network training and real-world deployment is available at https://github.com/IMRL/OriField.
\end{abstract}


\section{Introduction}
\label{sec:intro}
\begin{figure}[t]
    \centering
    \includegraphics[width=0.98\linewidth]{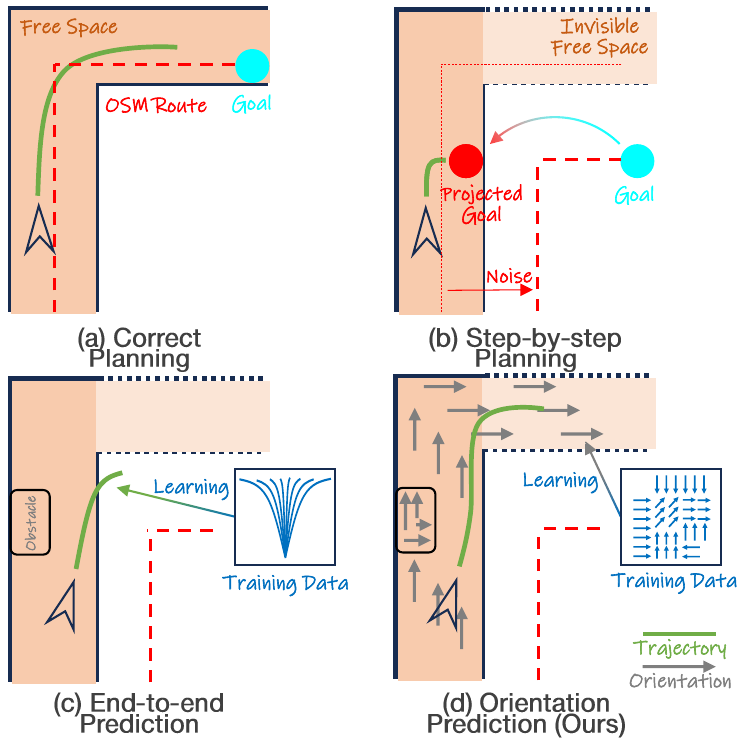}
    \caption{The comparison of our method with the previous step-by-step and end-to-end approach. (a): The correct planning result when observations and OSM routes contain minimal noise. (b): The step-by-step planning result caused by incomplete observations and noisy OSM routes. (c): The end-to-end approach predicts an infeasible trajectory drawn from the distribution of training trajectories. (d): Our method plans the correct trajectory based on orientations (gray arrowed) predicted by the deep network. In all figures, the red dashed line represents the OSM route, and the green line represents the planned trajectory. 
    }
    \label{fig:architecture}
\end{figure}

\begin{figure*}[t]
    \centering
    \includegraphics[width=0.98\linewidth]{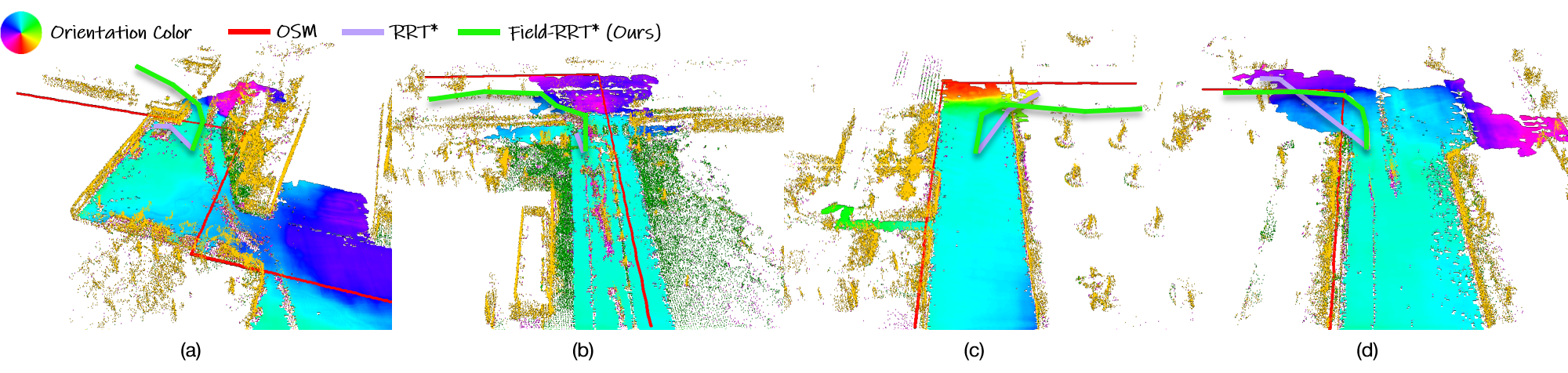}
    \caption{The visualization of the trajectory planning results of the typical implementation of the RRT* planner (purple) and our Field-RRT* planner (green). The red trajectory represents the OSM trajectory. Only the orientations within the free space are shown for clarity.}
    \label{fig:cover}
\end{figure*}
Robust trajectory planning is essential to ensure the safety of navigation systems. Most existing trajectory planning works are performed on prebuilt high-definition (HD) maps~\cite{bao2023review,zeng2019end,sadat2020perceive}, which provide precise and detailed environmental information. 
However, creating and maintaining HD maps requires significant resources and effort, particularly due to frequent changes in dynamic street scenes caused by road maintenance or adverse weather. These challenges make HD-map-based trajectory planning less practical for widespread and scalable deployment.

OSM\cite{haklay2008openstreetmap}-guided autonomous navigation without HD maps has gained considerable attention as a promising alternative~\cite{hentschel2010autonomous,floros2013openstreetslam} due to OSM's public accessibility and broader coverage of geographical areas. However, OSM data are inherently noisy and less detailed than HD maps, and it is crucial to establish the correspondence between OSM and local perception. To achieve this, most existing OSM-guided methods attempt to plan trajectories by leveraging additional sensors and data sources, such as LiDAR or GPS~\cite{suger2017global,ort2018autonomous,ort2019maplite,omama2023alt}. 
These methods often rely on registering the OSM route or optimizing a predefined trajectory model based on current observations, such as free space estimated from LiDAR scans.
 For example, some works \cite{ort2019maplite,omama2023alt} integrate segmentation, registration, global path planning, and local trajectory generation to enable navigation using simple OSM data. While these methods are effective when observations and OSM routes contain minimal noise (Fig. \ref{fig:architecture} (a)), they struggle under conditions of incomplete observations and noisy OSM routes, which are common due to the limited range of sensors, as depicted in Fig. \ref{fig:architecture} (b). As a result, they cannot handle corner cases, such as intersections, turns, or roads with ambiguous orientation. 
As a remedy, some works employ sliding windows or probability fusion to achieve comprehensive and reliable understanding by leveraging temporal information~\cite{suger2017global,ort2018autonomous}.
However, relying on the past time steps inherently faces short-horizon problems due to limited sensor range, and the performance is constrained by the accumulation of errors over successive steps. Moreover, registration and segmentation by themselves are difficult and inaccurate.

To address the above limitations, deep learning based methods plan trajectories in an end-to-end manner \cite{xu2022trajectory,paz2021tridentnet,paz2022tridentnetv2}, bypassing the need for explicit free space segmentation and achieving notable performance. However, these end-to-end methods rely heavily on high-quality training labels and lack scene understanding and explainability. 
Trained on limited data, these models inherit strong biases \cite{guo2022end} from human driving habits and are difficult to generalize effectively beyond the scope of the training data (Fig. \ref{fig:architecture} (c)). 

To overcome these challenges, we introduce a robust and explainable two-stage framework for trajectory planning, leveraging noisy OSM data for guidance. Specifically, we propose a novel representation called Orientation Field (OrField) tailored for trajectory planning based on which we can estimate the trajectory in the second stage through a trajectory planner. The key strength of OrField lies in its ability to generate a predictive field that guides the navigation of autonomous vehicles by integrating road geometry from LiDAR data and orientation preferences from OSM. To achieve this, we propose a data-driven approach with dedicated training data to encode both the initial OrField and LiDAR data. During the inference, our model takes the current LiDAR scan and OSM as input and generates a robust OrField without explicit registration and segmentation. 
As illustrated in Fig. \ref{fig:architecture} (d), the generated orientations (gray arrows) provide a clear and effective route even in invisible area.

Based on the estimated OrField, we propose two trajectory planners called Field-RRT* and Field-Bezier to plan the local trajectories by minimizing their deviation from the OrField (Fig. \ref{fig:cover}). This approach enhances the accuracy and robustness of generated trajectories, particularly in scenarios where traditional OSM-based methods fail. While the OrField is predicted using a data-driven method, our two-stage trajectory planning framework remains highly explainable, as it explicitly visualizes the orientations from any position on the map, even when dealing with noisy OSM data.
In summary, our contributions include:
\begin{itemize}
\item We propose a robust and explainable two-stage framework for trajectory planning with noisy OSM guidance.

\item We propose OrField, a novel data-driven orientation field that encodes both LiDAR observation and OSM information to provide a more comprehensive and accurate understanding of the road layout, improving the accuracy of trajectory planning.

\item Based on OrField, we develop two effective trajectory planners that estimate precise trajectories for autonomous navigation. Our planners excel in generating accurate paths for both straight and turning road segments, enhancing the robustness of autonomous navigation under OSM guidance.
\end{itemize}

\section{Related Works}
\subsection{Trajectory Planning Using Topometric Maps}
In contrast to the trajectory planning works relying on High Definition (HD) maps for autonomous driving \cite{cui2019multimodal,chai2019multipath, phan2020covernet}, recent works~\cite{suger2017global,ort2018autonomous,ort2019maplite,li2021openstreetmap,omama2023alt} attempted to accomplish the navigation task by only relying on topometric maps. For instance, \cite{suger2017global} utilized semantic terrain information from 3D-LiDAR data and the Markov-Chain Monte Carlo technique to align the OSM road route with the actual road center lines to obtain local navigation waypoints. However, this approach is limited by the accuracy of terrain semantic segmentation and struggles in complex road networks. \cite{ort2018autonomous} located the center of the road by extracting the road edges and employs RANSAC~\cite{bolles1981ransac} to fit the local trajectory. 
Similarly, \cite{li2021openstreetmap} optimized the local waypoint by extracting the center line of the road as a reference. A common limitation of these methods is their inability to effectively integrate OSM data for generating local trajectories in scenarios with ambiguous road boundaries or complex intersections. \cite{ort2019maplite} improved this situation by proposing a topometric map registration algorithm that approximates the maximum posterior estimate of the registration between the vehicle and the prior map, demonstrating its feasibility on CARLA road scenes \cite{dosovitskiy2017carla} with complex topologies. To further enhance localization in the road network, ALT-Pilot \cite{omama2023alt} combined LiDAR and camera cues with semantic information from OSM maps, generating local trajectories using the Frenet planner \cite{werling2010optimal}. ~\cite{tsiakas2023leveraging} focused on segmenting precise free spaces with LiDAR and camera to establish reliable local trajectory targets. 
Unlike these methods, which rely on complex algorithmic rules, we propose a novel data-driven representation that directly encodes both LiDAR scans and OSM data without the need for explicit registration and segmentation, significantly enhancing performance across various scenarios.

\subsection{Trajectory Planning via End-to-End Learning}
End-to-end learning approaches have emerged as a powerful paradigm for trajectory planning in navigation, enabling models to learn directly from raw sensor inputs and generate trajectories without the need for explicit registration and segmentation.  
Early works utilized CNNs to model spatial and temporal dynamics~\cite{zeng2019end}, but these methods often failed with complex real-world scenarios such as intersections and occlusions. To address these challenges, subsequent methods incorporated attention mechanisms~\cite{gu2023vip3d} and graph-based models~\cite{zhao2021tnt,gu2021densetnt,liu2024laformer} to better capture spatial-temporal dependencies. 
While these methods relied heavily on HDMaps for feature extraction \cite{bao2023review,sadat2020perceive}, Transfuser~\cite{chitta2022transfuser} and iPlanner~\cite{yang2023iplanner} generated the path given the local target without relying on a predefined HDMap.
Recently, there has been growing interest in utilizing simple map priors. 
Methods like multi-modal prediction~\cite{chai2019multipath,cui2019multimodal,phan2020covernet,deo2020trajectory,guo2022end} can be employed for this purpose as the task of trajectory prediction for the ego vehicle. 
\cite{xu2022trajectory,paz2021tridentnet} adopted bird's-eye-view (BEV) representations to encode navigation routes using rasterized maps. Tridentnetv2~\cite{paz2022tridentnetv2} advanced these representations by integrating graphical information for route encoding.
Learning from expert data with limited size and optimality, these methods converge toward sub-optimal solutions. \cite{shah2023gnm} uses 60 hours of video data including trajectory information from multiple robots and environments to improve generalization.
However, challenges such as generalization, explainability, and scalability persist, especially in rare or unseen conditions that have yet to be solved, which motivates us to propose a more explainable and robust framework to ensure accurate and reliable trajectory planning in complex environments.

\begin{figure*}[!ht]
    \centering
    \includegraphics[width=0.8\linewidth]{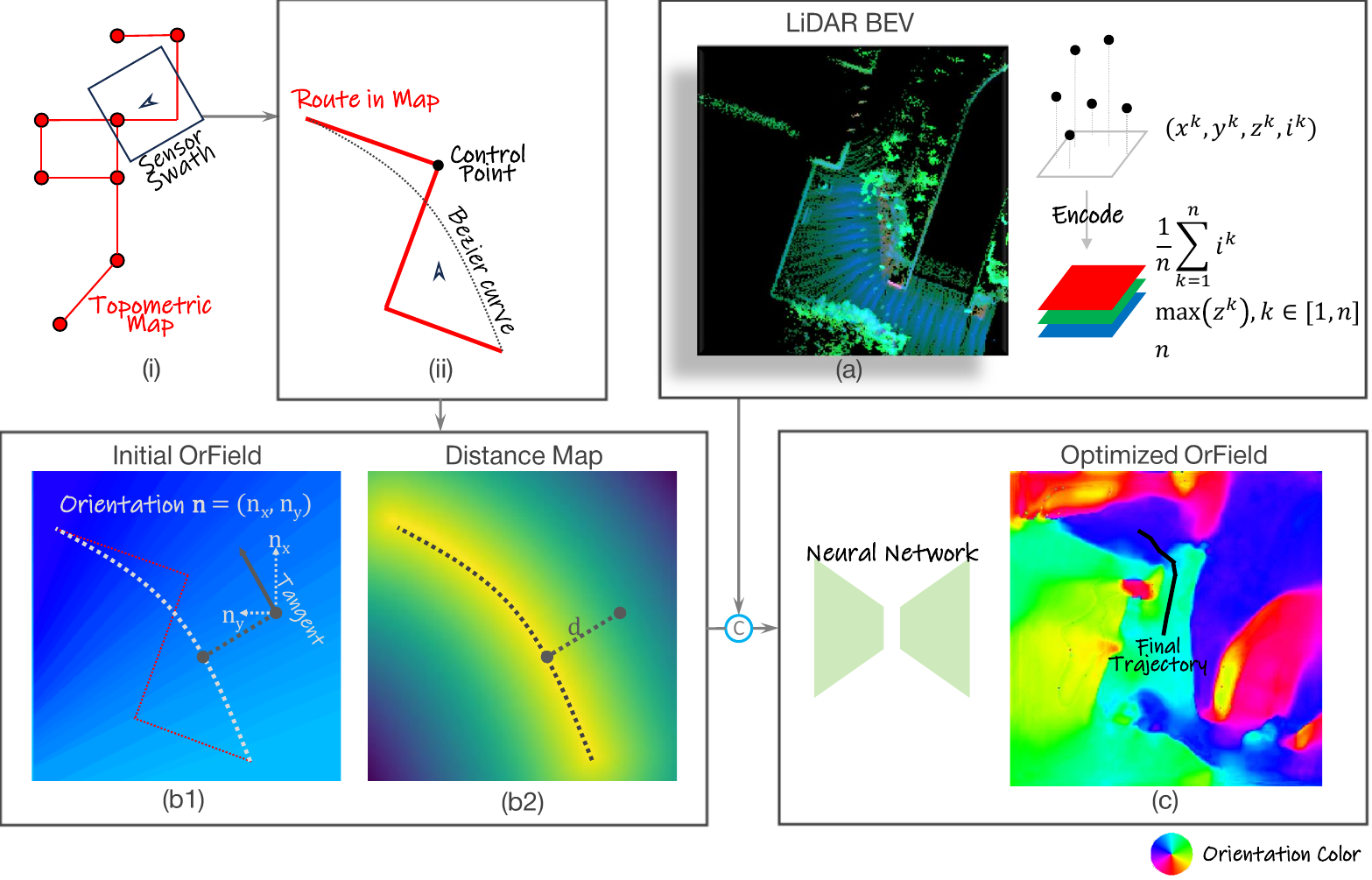}
    \caption{The examples of the LiDAR BEV, the initial OrField, the distance map, and the optimized OrField.
    (i): The global topometric map.
    (ii): The navigation route in sensor swath and its corresponding Bezier curve.
    (a): An example of BEV grid map generated by LiDAR scan with the average intensity, maximum height, and the number of points as features (colored in red, green, and blue, respectively). 
    (b1): The initial OrField at the intersection of (a). The orientation vector $\textbf{n}$ is decided by the tangent of the closest point on the Bezier curve generated by the navigation route.
    (b2): The distance map at the intersection of (a). The distance is the distance to the closest point on the Bezier curve.
    (c): The optimized OrField optimized by our trained deep network.}
    \label{fig:basic}
\end{figure*}

\section{Methodology}
We plan the trajectory in two stages: OrField and trajectory estimation.
First, we estimate the smooth initial OrField from coarse OSM directly, which is then optimized through a deep network with the aid of LiDAR scans.
In the second stage, we propose field planners to estimate the local trajectories by minimizing their orientation deviation from the OrField.

\subsection{Initial OrField}
\label{sec:ngm}
Our OrField can provide an orientation vector at each grid of the 2D space, serving as guidance for robot navigation, shown by the orientation color on the ground in Fig. \ref{fig:cover}. The estimation starts with an initialization of the OrField. We begin by rasterizing the environment into bird's eye view (BEV) grids. Each grid is assigned an orientation vector $\mathbf{n} = ( n_x, n_y )$, where $n_x$ and $n_y$ denote the orientation components along the x- and y-axes, respectively, as illustrated in Fig. \ref{fig:basic} (b1). The vector $\mathbf{n}$ defines the direction in which a robot is encouraged to move within the grid, and the length
$\|\mathbf{n}\|_2 \in [0, 1]$ quantifies the strength of the movement. A higher value of $\|\mathbf{n}\|_2$
  indicates a stronger preference for traversing that grid cell. In a local vicinity, the OrField can be utilized to plan a trajectory for the robot by iteratively following the suggested orientations. The OrField is thus an essential tool for guiding robot navigation over short distances, where the robot can optimize its trajectory by adhering to the directions encoded in the field.

To generate the initial OrField, we utilize the waypoints derived from the OSM's GPS coordinates, as illustrated in Fig. \ref{fig:basic} (i) and (ii). These waypoints are extracted from the topometric map of OSM in the sensor's swath, and a local route can be formed by connecting these waypoints. However, these paths are generally coarse and non-smooth making the network training difficult. To facilitate the network training, we turn these routes into Bezier curves, shown as the dark dashed curves in Fig. \ref{fig:basic} (ii), to produce a continuous and differentiable path. The endpoints of the Bezier curve coincide with the waypoints on the route and the control points are selected at the locations with the maximum offset from the auxiliary line connecting the two endpoints.

Once the Bezier curve is generated, the orientation of each grid cell in the OrField is initialized by the tangent at the closest point on the curve, as illustrated in Fig. \ref{fig:basic} (b1). Since this OrField is derived solely from the noisy OSM coordinates and does not consider the actual free space and obstacles, it is referred to as the initial OrField. As a result, the initial OrField is inherently coarse and may lead to inaccuracies, particularly at intersections or sharp turns. Consequently, the orientation vectors in these areas are not always ideal for robot navigation. To mitigate these issues, an optimization scheme is essential for the initial OrField.

\subsection{Optimized OrField}
\label{sec:deepngm}
To optimize the initial OrField, we propose leveraging LiDAR to help with road navigation and the avoidance of obstacles. A straightforward approach using LiDAR is to estimate the free space from current LiDAR observation, followed by the registration with the OSM routes to address the limitations of relying solely on OSM data. However, this approach still faces several challenges. Discontinuities in free space estimation, caused by occlusions or wrong segmentation, 
can render difficult path generation because of incorrect understanding of road structure, as shown in Fig. \ref{fig:architecture} (b). 
Such issues often arise when large occlusions lead to incomplete sensor observations. Additionally, the registration process is inherently complex, adding additional difficulty to the optimization pipeline.
To overcome these challenges, we optimize the initial OrField using a deep neural network that can accurately model the mappings between the initial OrField and the optimized OrField. 

\subsubsection{Training Data}
The network inputs include LiDAR BEV, initial OrField, and a distance map, while the output is the optimized OrField supervised with the orientation label we generated. 

\textbf{LiDAR BEV}. Following \cite{xu2022trajectory}, LiDAR measurements are represented in BEV with each grid equipped with three features: average intensity, maximum height, and number of points (Fig. \ref{fig:basic} (a)). 
During inference, partial observation is utilized only comprising past and current frames because future frames are unknown. The OrField is expected to be more adhere to the actual free space even with the aid of LiDAR BEV derived from only partial observation. Thus, during the training stage, the LiDAR BEV for network input is partial observation without aggregating future frames.
On the contrary, full observation of the LiDAR scans comprising past, current, and future frames is able to ensure precise free space detection, facilitating the generation of training data. Based on the full observation of scans, we generate versatile initial OrFields with distance maps and accurate orientation labels for network training, which are introduced below. 

\begin{figure*}[t]
    \centering
    \includegraphics[width=1.0\linewidth]{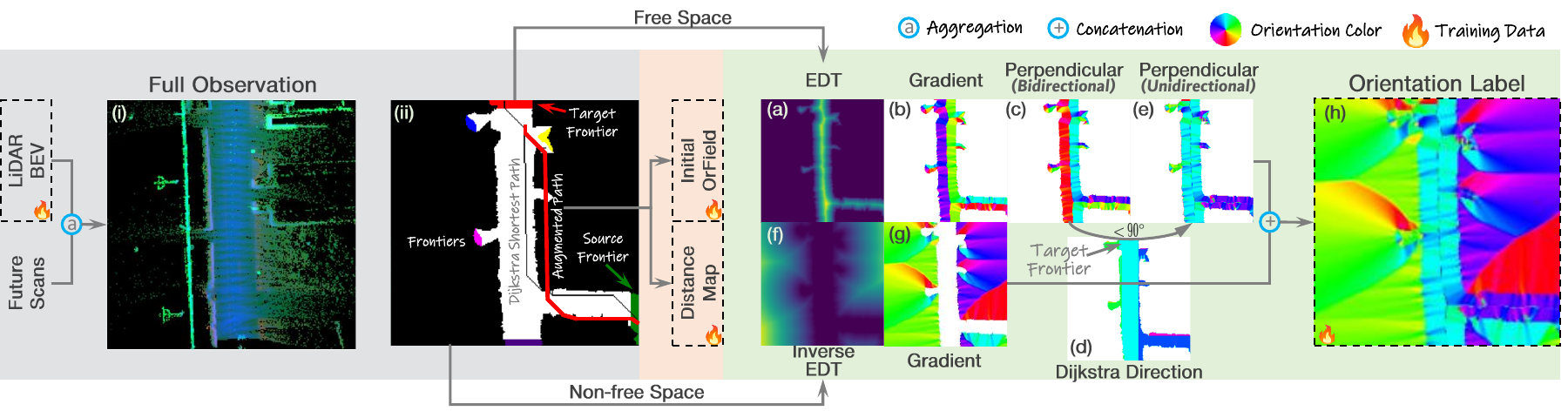}
    \caption{The training sample generation. A training pair consists of an initial OrField and a orientation label generated from a full observation aggregating past, current, and future frames.
    (i) The BEV of the point cloud aggregated by the consecutive past, current, and future LiDAR scans with odometry information. 
    (ii) The free space, frontiers, and the Dijkstra shortest path connecting the selected source and target frontiers in (i).
    At the right, it shows the Euclidean distance transform (EDT) and orientation calculation based on the EDT. (a) the EDT of the free space. (b) the gradient direction of (a). (c) the direction perpendicular to (b). (d) the Dijkstra shortest path tree generated from the target frontier. (e) the direction, both perpendicular to (b) and coherent with the direction of (d). (f) the inverse EDT of the free space. (g) the gradient direction of (f). (h) the combination of (e) and (g).}
    \label{fig:pipeline}
\end{figure*}

\textbf{Initial OrField}. 
We employ LiDAR segment methods to segment and extract the free space of full observations. Once the free space is identified, a pair of road frontiers is selected, representing the source and target points, as illustrated in Fig. \ref{fig:pipeline} (ii). The shortest path connecting these frontiers is computed using the Dijkstra algorithm. This path is then augmented with random shifts to generate augmented path, as depicted in Fig. \ref{fig:pipeline} (ii), which are subsequently transformed into Bezier curves. Finally, the initial orientation for each grid is determined by assigning the tangent vector of its closest point on the Bezier curve, as shown in Fig. \ref{fig:basic} (b1). This process ensures the creation of high-quality initial OrFields tailored for subsequent optimization. 

\textbf{Distance Map}. 
The distance map is constructed by calculating the distance from each BEV grid to the Bezier curve, as shown in Fig. \ref{fig:basic} (b2), to represent orientation confidence within each grid. Grids closer to the Bezier curve are assigned higher confidence values. Although the distance map is derived from OSM data during inference, it provides a flexible way to rectify the OrField, allowing effectively use of the inaccurate OSM data. Unlike the binary nature of OSM data, which categorizes grid cells strictly as either on or off the route (1 or 0), the distance map offers a continuous soft mapping that enables more adaptable adjustments. The effectiveness of the distance map is shown by ablation studies in Section \ref{sec:ablations2}.

\textbf{Orientation Label}. 
\label{sec:optimizedfiled}
The OSM is inaccurate by nature and our training target is to estimate accurate OrField to plan the driving trajectory. To achieve this, we generate orientation labels for network training based on the actual road structure. This process begins by computing the Euclidean Distance Transformation (EDT), which represents distances to the nearest road border on the free space of the full observations (Fig. \ref{fig:pipeline} (a)), and the inverse EDT (Fig. \ref{fig:pipeline} (f)). Next, the gradient direction of the EDT (Fig. \ref{fig:pipeline} (b)) is calculated, typically pointing perpendicular to the closest road boundary. A second gradient operation is applied to derive the direction parallel to the road border (Fig. \ref{fig:pipeline} (c)), though this orientation is ambiguous, as it may align with or oppose the preferred travel direction. To resolve this ambiguity, we calculate the Dijkstra direction (Fig. \ref{fig:pipeline} (d)) using the Dijkstra algorithm on the free space from the target frontier. The preferred orientation is determined by ensuring its alignment with the Dijkstra path direction, validated by a positive dot product between the two vectors. For grid cells outside the free space, the negative gradient of the inverse EDT (Fig. \ref{fig:pipeline} (g)) is used as the preferred orientation, encouraging movement toward the free space from obstacles. Finally, the orientations on the free space and those outside it are combined to produce the orientation label for training (Fig. \ref{fig:pipeline} (h)). This approach ensures the robot's movement aligns with the road structure while achieving obstacle avoidance and effective navigation.

\begin{figure*}[t]
    \centering
    \includegraphics[width=1.0\linewidth]{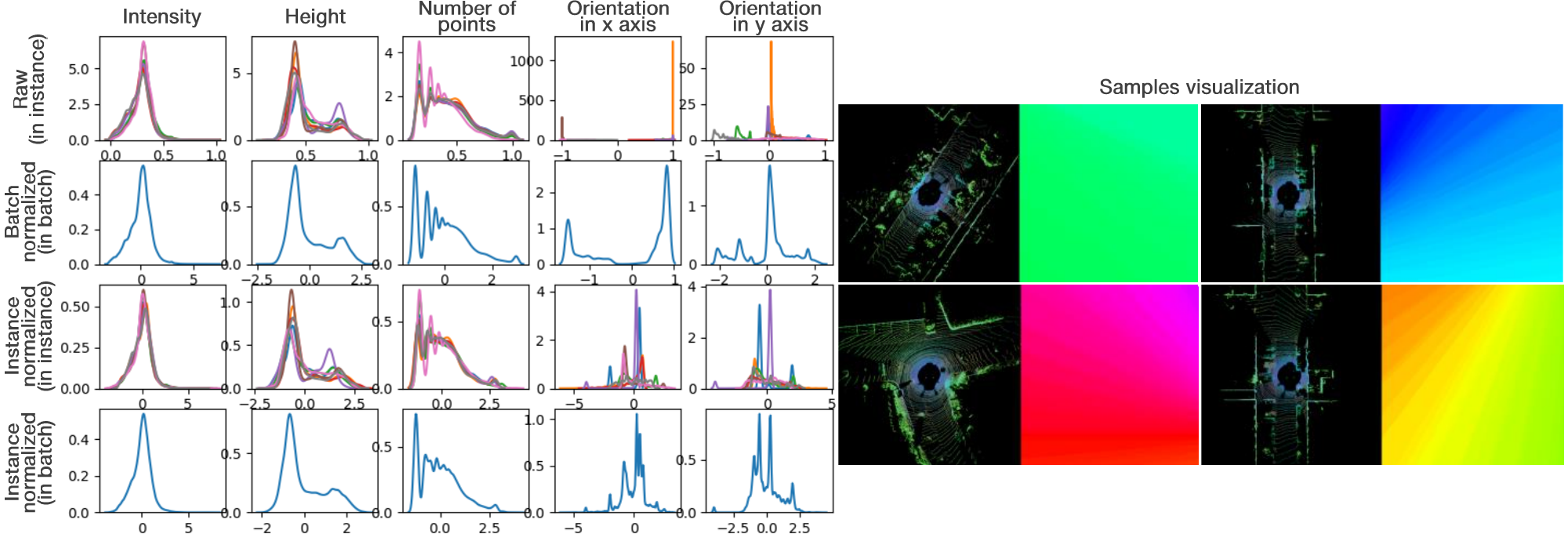}
    \caption{An example of a training batch with 8 samples. On the left are the feature distributions of this batch, and on the right is the visualization of 4 examples in this batch. \textbf{Row 1 on the left:} Distributions of 5 features for network inputs. From the left to the right are the average intensity, the maximum height, the number of points, the direction along the x-axis, and the direction along the y-axis. \textbf{Row 2 on the left:} The batch-normalized distributions of these features. \textbf{Row 3 on the left:} The instance-normalized distributions of these features. Samples are shown in different distributions. \textbf{Row 4 on the left:} The instance-normalized distributions of these features. Samples are shown in one distribution. \textbf{Right:} The visualization of 4 examples in this batch.}
    \label{fig:divergesample}
\end{figure*}

\subsubsection{Network Architecture}
We adopt the SalsaNext architecture \cite{cortinhal2020salsanext}, a highly effective network for processing sparse LiDAR data using dilated convolutions \cite{YuKoltun2016}. While the original SalsaNext is designed to process LiDAR range data, we modify it to accept LiDAR BEV grids, the initial OrField, and the distance map as inputs, producing an optimized OrField as output. As a result, the data normalization module also needs modification. In the original SalsaNext, batch normalization stabilizes training by assuming that input features, such as normalized x, y, and z coordinates, range, and intensity, follow Gaussian-like distributions with similar mean and variance. However, our inputs, particularly orientation data, exhibit significant variability across samples, as shown in Fig. \ref{fig:divergesample}. This variability causes inconsistent outputs during training and a distribution shift between training and inference, which undermines generalization. To address these challenges, we use instance normalization, which normalizes each sample independently. This ensures stable distributions during training and consistent behavior between training and inference, regardless of batch size. This modification enhances the network's robustness and adaptability to our application.

\subsubsection{Loss Function}
We train the network by minimizing the difference between predicted and target orientations in polar coordinates. For each grid, the target orientation vector $\mathbf{n} = ( n_x, n_y )$ from the orientation label is converted into a corresponding angle $\theta_n \in [-\pi, \pi]$ using $\theta_n = \text{arctan2}(n_y, n_x)$. Rather than directly predicting the orientation in $[-\pi, \pi]$, we train the network to predict a radian offset $\Delta\theta \in \mathbb{R}$ that adjusts the initial orientation vector $\mathbf{d}$ from the initial OrField to the orientation label. The difference $r$ between the predicted and ground truth orientations is computed as $r = (\theta_n - (\theta_d + \Delta\theta)) \mod (2\pi)$, where $\theta_d$ is the corresponding angle of orientation vector $\mathbf{d}$. This normalized difference $r \in [0, 2\pi]$ is interpreted such that values in the interval $[0, \pi]$ indicate increasing differences, while those in $[\pi, 2\pi]$ indicate decreasing differences. To map $r$ to the interval $[-\pi, \pi]$, we define $l = r - 2\pi$ when $r > \pi$, and $l = r$ when $r \leq \pi$. The final loss function is then given by: 
$loss = \sum_i \left| l_i \right|$
where $\left| l_i \right|$ is the angular difference for grid $i$.
During inference, the initial orientation vector $\mathbf{d}$ is adjusted by the predicted offset $\Delta\theta$. The predicted orientation components along the x- and y-axes are then computed as $\hat{n_x} = \cos(\theta_d + \Delta\theta)$ and $\hat{n_y} = \sin(\theta_d + \Delta\theta)$.

\begin{figure*}
    \centering
    \includegraphics[width=0.88\linewidth]{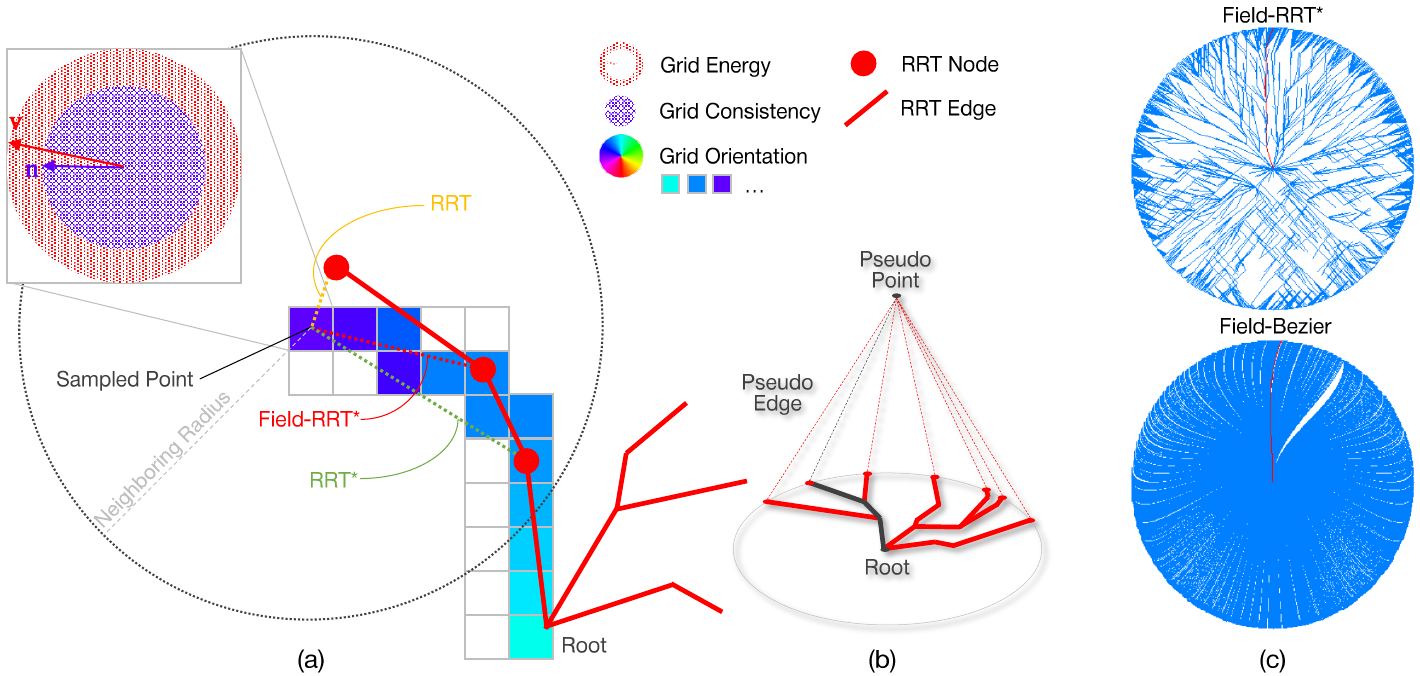}
    \caption{
    \textbf{(a)} The illustration of tree growing. 
    A newly sampled point selects its parent among its neighbors with the lowest Euclidean distance (yellow), total Euclidean distance (green), and total grid energy $E$ (red). 
    The colored grids show the orientation $\mathbf{n}$ (Section \ref{sec:ngm}) along the path. On the zoom-in of each grid, the red areas show the energy $e$ of this grid. The blue and red arrow lines are the orientation $\mathbf{n}$ at this grid and the normalized orientation $\mathbf{v} \in \mathbb{R}^2$, $\|\mathbf{v}\|_2 = 1$ of the edge cross this grid, respectively. 
    \textbf{(b)} The illustration of target point. 
    The solid lines show the RRT edge with energy defined in (a), while the dashed lines show the pseudo edge with energy defined as $0$. The black points show the tree root and the pseudo point. The red points show the RRT leaves. The black path is the one with the lowest total energy $E$ to reach the target point.
    \textbf{(c)} The comparison between Field-RRT* and Field-Bezier. The candidate paths are shown in blue. The final path is shown in red. Top: Field-RRT*. Bottom: Field-Bezier.
    }
    \label{fig:problem}
\end{figure*}

\subsection{Field Planner}
Based on the optimized OrField, we propose two field planners, \textbf{Field-RRT*} and \textbf{Field-Bezier}, respectively, to estimate the navigation trajectory. 

\textbf{Field-RRT*.}
Our approach is based on RRT*, an extension of the RRT algorithm, which finds a path to a target point by iteratively sampling new points, expanding the tree from these points, and eventually reaching the target through the tree's leaves. Each time sampling a point, the algorithm reconnects the edges to the neighboring nodes, ensuring that the path is progressively shortened with each iteration. However, since the neighbor points are typically identified based on Euclidean distance, the resulting path may not align well with the OrField.

To address this issue, we propose Field RRT*, where the tree is grown under the constraint of the OrField. The demostration is shown in Fig. \ref{fig:problem} (a). Specifically, when a new point is sampled, the algorithm selects the point with the lowest energy $E$ from the root to the sampled point (linked via the red dotted line). The total energy $E$ is the sum of the energy values $e$ of all grids along the path from the tree root to the sampled point. For each grid, $e$ is the area of the red region in the unit circle (as shown in the zoom-in of the grid in Fig. \ref{fig:problem} (a)), determined by the difference between the orientation vector $\mathbf{n}$  (Section \ref{sec:ngm}) on that grid and the normalized orientation vector $\mathbf{v} \in \mathbb{R}^2$, $\|\mathbf{v}\|_2 = 1$, representing the direction of the edge crossing the grid.
For the orientations optimized by our network, $\|\mathbf{n}\|_2$ is equal to $1$. We merge the orientation of each grid with its neighbors using a 2-dimensional uniform filter, producing a shortened $\mathbf{n}$ satisfying $\|\mathbf{n}\|_2 \leq 1$ from which the inequality holds when the orientations of neighboring grids are inconsistent.
After selecting the parent, the algorithm searches for neighboring nodes and attempts to reconnect them as children, ensuring that the total energy $E$ is minimized. As shown in Fig. \ref{fig:problem} (a), the path generated by Field RRT* better aligns with the actual orientation, resulting in a path that adheres more closely to the desired trajectory.

To mimic the RRT-like algorithms, let us assume there exists a target point, which can be virtually a pseudo target floating in the sky. Given the definition of distance from the RRT leaves to the pseudo target (as illustrated in Fig. \ref{fig:problem} (b)), the tree grows similarly to the RRT-like algorithms which find the path with the smallest distance from the root to the target point. Different from the existing RRT-like approaches, we find a path with the smallest deviation from the predefined OrField. 
In this way, the path is optimized over iterations. The algorithm terminates after a predefined number of iterations.

\textbf{Field-Bezier.}
We propose Field-Bezier by introducing driving constraints to ease trajectory planning. Specifically, a batch of Bezier curves are generated from the vehicle to the points on the circle center at the vehicle with a radius equal to a given planning distance. The orientation of the start and end points of the Bezier curve are the same as in the given field. The first and second control points of the Bezier curve are generated by moving the start point forward along its direction and moving the endpoint backward against its direction for a fixed distance, respectively. The Bezier curve is generated by the start and end points, and the two control points. The points and tangents of the generated Bezier curves are sampled at the grids that are occupied by the curve. The energy of each generated Bezier curve is calculated in a similar way to the Field-RRT* algorithm. Finally, we choose the Bezier curve with the lowest energy as the planned trajectory. The comparison between Field-RRT* and Field-Bezier is shown in Fig. \ref{fig:problem} (c).

\section{Experiments}
In this section, we conduct experiments on two datasets including a public SemanticKITTI dataset~\cite{behley2019semantickitti} and one dataset collected on our campus. We select the latest end-to-end trajectory prediction work using topometric map~\cite{xu2022trajectory} as the end-to-end baseline and a segmentation-based method that plans trajectories based on segmented free space \cite{gao2024active} as segmentation baseline. 
The evaluation includes frame level and trajectory level. At the frame level, metrics are designed to compare the differences between the planned trajectory and the ground truth trajectory frame-by-frame. At the trajectory level, all the planned trajectories from each frame are temporally aggregated, and the quality of the aggregated trajectory is accessed. 
To comprehensively evaluate performance in real-world scenarios, we deployed both the end-to-end baseline and ours on a ground vehicle equipped with an Intel NUC 11 computer. The assessment involves comparing the number of manual interventions if required and the deviation between the vehicle's trajectory and the road centerline, which is defined and recorded by manual driving routes.

\subsection{Metrics}
\begin{figure}
    \centering
    \includegraphics[width=0.95\linewidth]{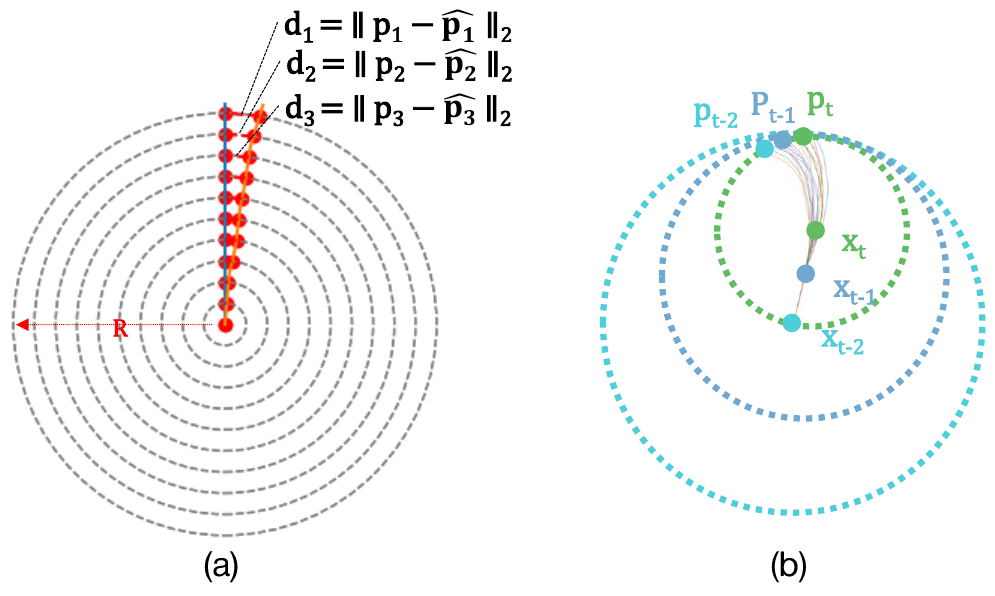}
    \caption{The illustration of trajectory sampling (a) and trajectory fusion (b).}
    \label{fig:metricpie}
\end{figure}

At the frame level, we use the widely used metrics, ADE, FDE, and HitRate, for evaluating trajectory planning performance.
The Average Displacement Error (\textbf{ADE}) measures the average difference between the planned trajectory and the ground truth trajectory generated from ground truth poses in datasets. It is defined as $\frac{1}{N}\sum_{i=1}^{N} \|p_i - \hat{p_i}\|_2$, where $p$ and $\hat{p}$ are the sampled points in the ground truth trajectory and planned trajectory. The Final Displacement Error (\textbf{FDE}) depicts the position difference between the last sampled points in the ground truth and the planned trajectory, respectively. It is defined as $\|p_{N} - \hat{p}_{N}\|_2$. The last sampled point can be used as the goal for local planning. 
The \textbf{HitRate} \cite{phan2020covernet} $\in \{0, 1\}$ is defined as whether the planned trajectory is hit. A hit happens if $\max_{i=1}^{N} \|p_i - \hat{p_i}\|_2 < d$ where $d$ is a distance threshold. In addition, we measure the \textbf{Coverage} of the planned trajectory, defined as $\frac{1}{N}\sum_{i=1}^{N} (\|p_i - \hat{p_i}\|_2 < d)$, which is a "soft" HitRate describing the average hit over all sampled points along the trajectory. The sampling method is shown in Fig. \ref{fig:metricpie} (a). Different sampling radii $R$ are used for comprehensive evaluations on trajectory performance. Finally, we compute the average of all these metrics over the entire trajectory.

At the trajectory level, the planned trajectory is temporally fused, as shown in Fig. \ref{fig:metricpie} (b). $x_{t-2}$, $x_{t-1}$, and $x_{t}$ are the positions of the past two time steps and the current time step, respectively. Each time step provides a planned trajectory with a waypoint, i.e., $p_{t-2}$, $p_{t-1}$, $p_{t}$, to the same distance head $x_{t}$. We fuse these predictions according to a prior prediction variance (empirical setting) and obtain the fused waypoint $\hat{p}_{t}$ for $x_{t}$. We later plot the fused waypoints of all the time steps $t$ and generate the entire planned trajectory over the scene for visualization.

\begin{table*}[htb!]
  \caption{Trajectory planning comparisons of the segmentation baseline (\textbf{Se.}), end-to-end baseline (\textbf{En.}), our method with Field-RRT*(\textbf{RRT.}) and Field-Bezier (\textbf{Bez.}) on SemanticKITTI dataset. All these metrics are evaluated from the current position to $10$ meters and $20$ meters ahead.}
  \label{tab:baselinesseqs}
  \renewcommand{\arraystretch}{1.0}
  \setlength{\tabcolsep}{4pt}
  \centering
  \begin{tabular}{@{}r|cccc|cccc|cccc|cccc}
\toprule
\multicolumn{1}{c}{\multirow{2}{*}{Seq}} & \multicolumn{1}{c}{\textbf{Se.}\cite{gao2024active}} & \multicolumn{1}{c}{\textbf{En.}\cite{xu2022trajectory}} & \multicolumn{1}{c}{\textbf{RRT.}} & \multicolumn{1}{c}{\textbf{Bez.}} & \multicolumn{1}{c}{\textbf{Se.}\cite{gao2024active}} & \multicolumn{1}{c}{\textbf{En.}\cite{xu2022trajectory}} & \multicolumn{1}{c}{\textbf{RRT.}} & \multicolumn{1}{c}{\textbf{Bez.}} & \multicolumn{1}{c}{\textbf{Se.}\cite{gao2024active}} & \multicolumn{1}{c}{\textbf{En.}\cite{xu2022trajectory}} & \multicolumn{1}{c}{\textbf{RRT.}} & \multicolumn{1}{c}{\textbf{Bez.}} & \multicolumn{1}{c}{\textbf{Se.}\cite{gao2024active}} & \multicolumn{1}{c}{\textbf{En.}\cite{xu2022trajectory}} & \multicolumn{1}{c}{\textbf{RRT.}} & \multicolumn{1}{c}{\textbf{Bez.}}\\
\cmidrule(r){2-5}\cmidrule(r){6-9}\cmidrule(r){10-13}\cmidrule(r){14-17}
\multicolumn{1}{c}{} & \multicolumn{4}{c}{ADE$_{10m}\downarrow$} & \multicolumn{4}{c}{FDE$_{10m}\downarrow$} & \multicolumn{4}{c}{HitRate$_{10m}\uparrow$} & \multicolumn{4}{c}{Coverage$_{10m}\uparrow$}\\
\cmidrule(r){2-5}\cmidrule(r){6-9}\cmidrule(r){10-13}\cmidrule(r){14-17}
08 & 0.65 & 0.27 & 0.25 & \textbf{0.21} & 0.97 & 0.44 & 0.51 & \textbf{0.39} & 0.91 & 0.82 & 0.96 & \textbf{0.96} & 0.94 & 0.95 & 0.98 & \textbf{0.98}\\
13 & 1.08 & 0.29 & 0.27 & \textbf{0.24} & 1.84 & 0.52 & 0.52 & \textbf{0.46} & 0.83 & 0.79 & \textbf{0.96} & 0.95 & 0.88 & 0.94 & 0.97 & \textbf{0.98}\\
15 & 0.55 & 0.29 & 0.26 & \textbf{0.22} & 0.82 & 0.49 & 0.47 & \textbf{0.39} & 0.95 & 0.81 & \textbf{1.00} & 0.98 & 0.97 & 0.96 & \textbf{1.00} & 0.99\\
16 & 0.56 & 0.21 & 0.21 & \textbf{0.19} & 0.91 & \textbf{0.31} & 0.46 & 0.37 & 0.89 & 0.82 & \textbf{0.98} & 0.97 & 0.93 & 0.96 & 0.99 & \textbf{0.99}\\
18 & 0.52 & 0.14 & 0.16 & \textbf{0.13} & 0.64 & \textbf{0.20} & 0.42 & 0.26 & 0.97 & 0.88 & \textbf{1.00} & 1.00 & 0.99 & 0.98 & \textbf{1.00} & 1.00\\
19 & 0.94 & 0.27 & 0.27 & \textbf{0.21} & 1.22 & 0.51 & 0.53 & \textbf{0.43} & 0.73 & 0.80 & 0.96 & \textbf{0.96} & 0.88 & 0.95 & 0.98 & \textbf{0.98}\\
\midrule
\multicolumn{1}{c}{} & \multicolumn{4}{c}{ADE$_{20m}\downarrow$} & \multicolumn{4}{c}{FDE$_{20m}\downarrow$} & \multicolumn{4}{c}{HitRate$_{20m}\uparrow$} & \multicolumn{4}{c}{Coverage$_{20m}\uparrow$}\\
\cmidrule(r){2-5}\cmidrule(r){6-9}\cmidrule(r){10-13}\cmidrule(r){14-17}
08 & 0.88 & 0.53 & 0.44 & \textbf{0.41} & 1.29 & 1.16 & 0.89 & \textbf{0.87} & 0.85 & 0.81 & 0.85 & \textbf{0.87} & 0.91 & 0.93 & 0.95 & \textbf{0.95}\\
13 & 1.78 & 0.70 & 0.44 & \textbf{0.43} & 3.10 & 1.76 & \textbf{0.81} & 0.81 & 0.73 & 0.77 & 0.89 & \textbf{0.90} & 0.83 & 0.91 & 0.95 & \textbf{0.95}\\
15 & 0.80 & 0.57 & 0.44 & \textbf{0.37} & 1.42 & 1.20 & 0.89 & \textbf{0.73} & 0.87 & 0.80 & 0.90 & \textbf{0.92} & 0.94 & 0.93 & \textbf{0.98} & 0.97\\
16 & 0.79 & 0.37 & 0.37 & \textbf{0.36} & 1.13 & 0.76 & \textbf{0.67} & 0.67 & 0.81 & 0.84 & 0.91 & \textbf{0.93} & 0.90 & 0.96 & 0.97 & \textbf{0.97}\\
18 & 0.64 & \textbf{0.24} & 0.38 & 0.30 & 0.93 & \textbf{0.51} & 0.89 & 0.67 & 0.94 & 0.91 & 0.94 & \textbf{0.98} & 0.98 & 0.99 & 0.99 & \textbf{0.99}\\
19 & 1.15 & 0.70 & 0.45 & \textbf{0.42} & 1.67 & 1.79 & \textbf{0.80} & 0.81 & 0.64 & 0.76 & 0.88 & \textbf{0.91} & 0.84 & 0.91 & 0.95 & \textbf{0.96}\\
\midrule
  \end{tabular}
\end{table*}

\begin{figure*}[ht]
    \centering
    \includegraphics[width=1\linewidth]{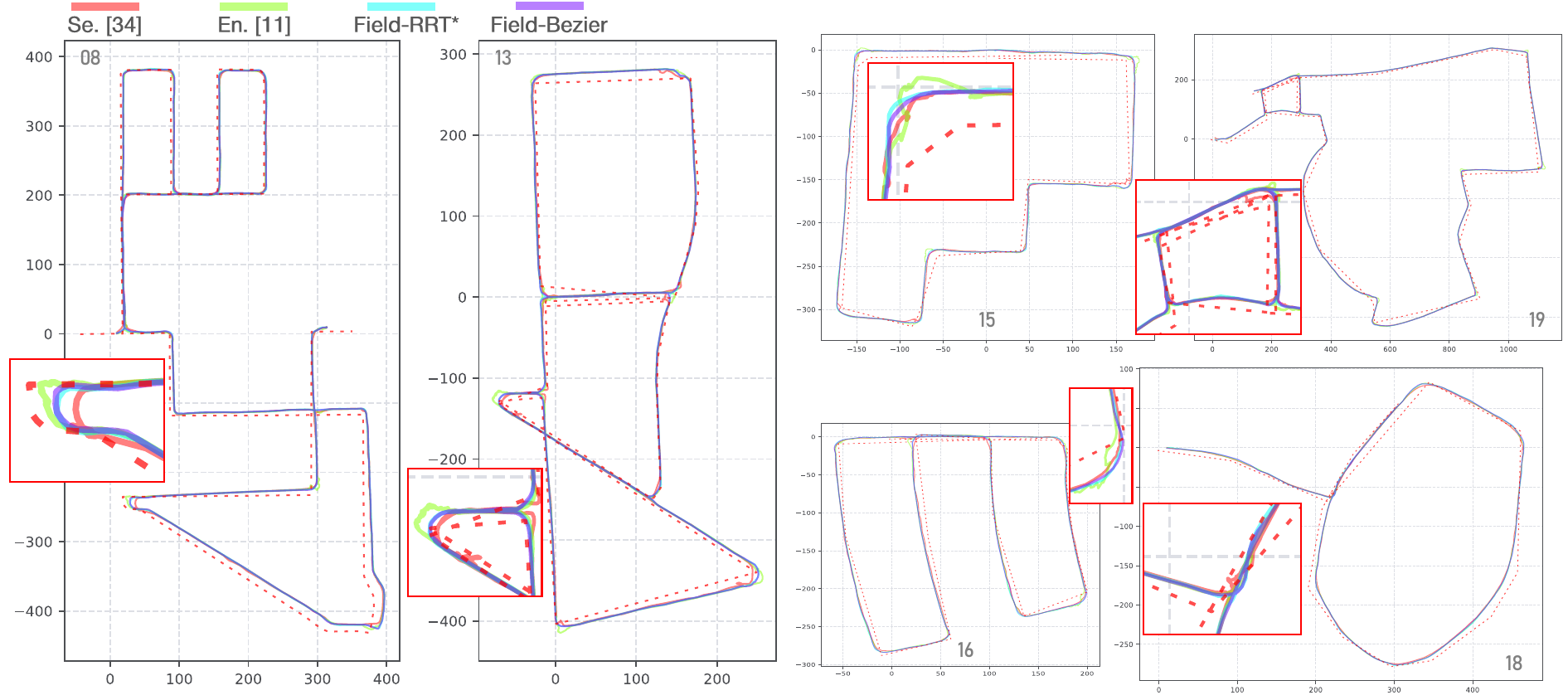}
    \caption{The planned waypoints along the trajectory (solid lines) and the OSM route (dashed line) on SemanticKITTI dataset. The baselines either deviate at the corner and intersection or shift at the straight road, while our method provides a seamless planning result along the trajectory.}
    \label{fig:baselinescmp}
\end{figure*}

\subsection{Results on SemanticKITTI}
SemanticKITTI \cite{behley2019semantickitti} and KITTI \cite{geiger2012we} are popular datasets for validating robotics and autonomous driving systems. Both datasets include LiDAR scans collected using the Velodyne-HDLE64 LiDAR. SemanticKITTI provides 22 sequences with semantic labels for each LiDAR point, with labels publicly available for sequences 00-10. We utilize KITTI for model training following \cite{xu2022trajectory} and evaluate performance in SemanticKITTI sequences. Both SemanticKITTI and KITTI include the ground truth poses, enabling the generation of ground truth trajectories for evaluation and straightforward implementation of relative pose estimation for point cloud aggregation. The step size, neighbor radius, and number of iterations of Field-RRT* are set to 1 m, 2 m, and 1000, respectively. The RRT tree is down-sampled half along the x- and y-axis for fast running.

To train our model, no OSM routes or manual driving trajectories are provided for supervision. The free space for training is estimated by the LiDAR segmentation method \cite{cortinhal2020salsanext}. 
We compare our method with the two baselines. To train the end-to-end baseline model, authors in \cite{xu2022trajectory} obtained OSM routes from the OSM website as input and ground-truth driving trajectories from the dataset collected by manual driving as supervision. We directly obtain the checkpoint released by the authors. The segmentation baseline uses the free space estimated by the same method used for generating training samples for our model.
Five SemanticKITTI sequences (sequences 08, 13, 15, 16, 18, and 19) are used for testing as they are in complex road topologies.
For evaluation, the OSM routes are downloaded from the OSM website and registered with the ground-truth driving trajectories.
During inference, these OSM routes are converted into initial OrField (Section \ref{sec:ngm}) for network input. 

Table \ref{tab:baselinesseqs} presents the quantitative results, where the Field-RRT* and Field-Bezier planners based on the learned OrField outperform the others in most sequences.
However, the end-to-end baseline achieves better for metric in distance error, i.e., ADE and FDE, when evaluate at $20$ meters in sequence 18 and evaluated at $10$ meters in sequences 16 and 18. This is because the end-to-end baseline trained the network overfits smooth driving, while sequences 16 and 18 are the two smooth driving sequences with the least frames categorized as sharp turns. This alignment between training and evaluation conditions results in stronger performance. Nonetheless, in real-world scenarios, where conditions are more diverse, an overfitted predictor may fail to generalize effectively. 

The qualitative comparison is shown in Fig. \ref{fig:baselinescmp}. The plotted trajectories are generated by fusing outputs from each time step, based on past frames. We observe that the end-to-end baseline trained with OSM guidance, as well as the segmentation baseline relying on heuristic designs, does not perform well in complex scenarios (especially at turns), as shown in the zoom-in at each sequence of Fig. \ref{fig:baselinescmp}. In contrast, our method with Field-RRT* and Field-Bezier planners demonstrates superior understanding in both straight-road and intersection scenarios, thanks to unbiased training and independent trajectory planning.
Notably, despite our network being trained without exposure to the realistic OSM route, i.e., the extent and structure of OSM routes are unknown during training, it generalizes effectively to real-world applications.

\begin{figure}[htb]
    \centering
    \includegraphics[width=1\linewidth]{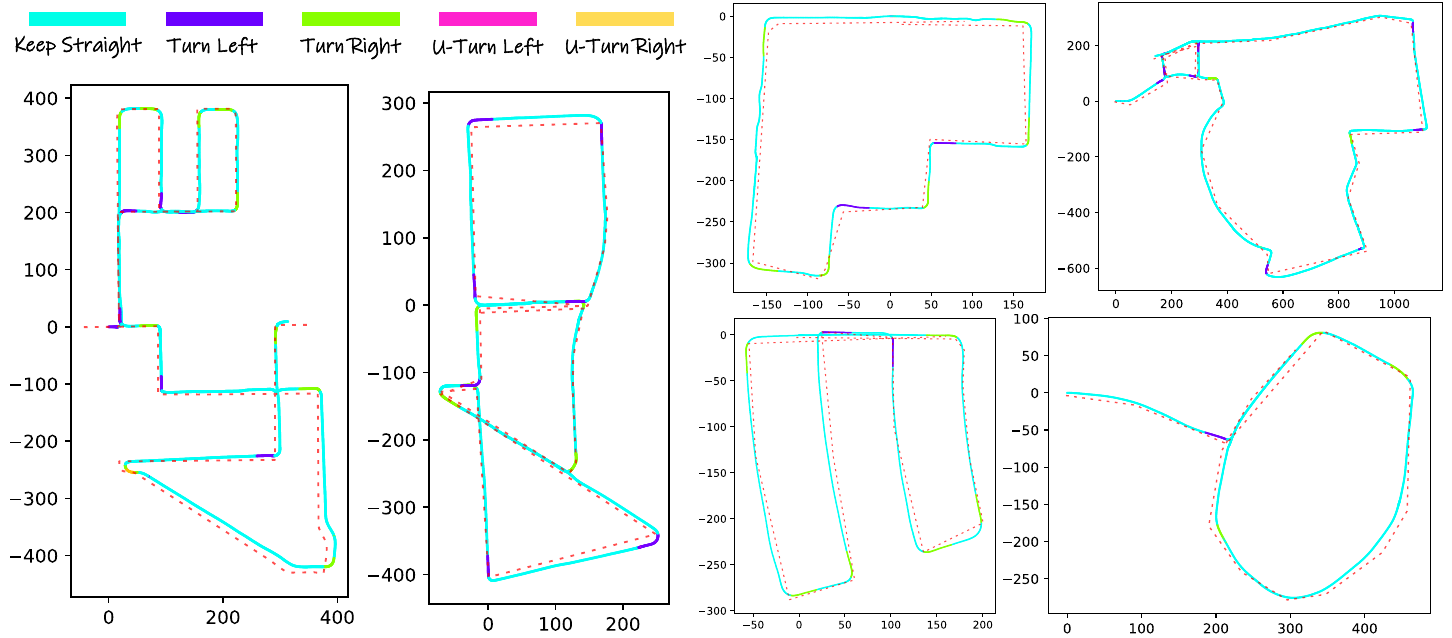}
    \caption{The 5 categories of vehicle directions on SemanticKITTI dataset. They are keep-straight, turn-left, turn-right, U-turn to the left, and U-turn to the right.}
    \label{fig:turntype}
\end{figure}

\begin{table}[ht]
    \caption{Comparisons of Field-RRT* (\textbf{RRT.}) and Field-Bezier (\textbf{Bez.}) with frames categorized by keep straight and turn on SemanticKITTI dataset. These metrics are evaluated from the current position to $20$ meters ahead.}
    \label{tab:previousart}
    \centering
    \renewcommand{\arraystretch}{1.0}
    \setlength{\tabcolsep}{8pt}
    \begin{tabular}{@{}rcc|cc|cc@{}}
    \toprule
    \multicolumn{1}{c}{\multirow{2}{*}{Seq}} & \multicolumn{1}{c}{\textbf{RRT.}} & \multicolumn{1}{c}{\textbf{Bez.}} & \multicolumn{1}{c}{\textbf{RRT.}} & \multicolumn{1}{c}{\textbf{Bez.}} & \multicolumn{1}{c}{\textbf{RRT.}} & \multicolumn{1}{c}{\textbf{Bez.}}\\
    \cmidrule(r){2-3}\cmidrule(r){4-5}\cmidrule(r){6-7}
    \multicolumn{1}{c}{} & \multicolumn{2}{c}{ADE$_{Straight}\downarrow$} & \multicolumn{2}{c}{FDE$_{Straight}\downarrow$} & \multicolumn{2}{c}{HitRate$_{Straight}\uparrow$}\\
    \cmidrule(r){2-3}\cmidrule(r){4-5}\cmidrule(r){6-7}
    08 & 0.31 & \textbf{0.26} & 0.58 & \textbf{0.53} & 0.93 & \textbf{0.95}\\
    13 & 0.28 & \textbf{0.27} & 0.53 & \textbf{0.51} & 0.97 & \textbf{0.98}\\
    15 & 0.39 & \textbf{0.29} & 0.73 & \textbf{0.52} & 0.95 & \textbf{0.99}\\
    16 & 0.29 & \textbf{0.26} & \textbf{0.52} & 0.55 & 0.97 & \textbf{0.99}\\
    18 & 0.36 & \textbf{0.27} & 0.80 & \textbf{0.56} & 0.95 & \textbf{1.00}\\
    19 & 0.38 & \textbf{0.34} & \textbf{0.69} & 0.72 & 0.92 & \textbf{0.95}\\
    \midrule
    \multicolumn{1}{c}{} & \multicolumn{2}{c}{ADE$_{Turn}\downarrow$} & \multicolumn{2}{c}{FDE$_{Turn}\downarrow$} & \multicolumn{2}{c}{HitRate$_{Turn}\uparrow$}\\
    \cmidrule(r){2-3}\cmidrule(r){4-5}\cmidrule(r){6-7}
    08 & \textbf{0.96} & 0.97 & \textbf{2.07} & 2.21 & \textbf{0.55} & 0.54\\
    13 & 0.97 & \textbf{0.95} & \textbf{1.75} & 1.82 & 0.60 & \textbf{0.65}\\
    15 & \textbf{0.60} & 0.65 & \textbf{1.41} & 1.44 & \textbf{0.74} & 0.73\\
    16 & \textbf{0.67} & 0.70 & 1.20 & \textbf{1.12} & 0.69 & \textbf{0.75}\\
    18 & \textbf{0.62} & 0.63 & \textbf{1.82} & 1.86 & 0.81 & \textbf{0.88}\\
    19 & 0.84 & \textbf{0.83} & 1.43 & \textbf{1.26} & \textbf{0.66} & 0.65\\
    \midrule
    \end{tabular}
\end{table}

\subsection{Field-RRT* versus Field-Bezier}
From Table~\ref{tab:baselinesseqs}, we see that Field-Bezier outperforms Field-RRT* for most metrics in most sequences. This observation motivates us to investigate the benefits of the behavior priors introduced by the trajectory planner. Field-RRT* imposes no prior knowledge, estimating trajectories purely based on orientations, whereas Field-Beier enforces strong constraints, restricting trajectories to Bezier curves. To analyze whether incorporating behavior priors benefits trajectory estimation, we classify the frames into "keep-straight" or "turn" scenarios by examining the direction suggested by the OSM trajectory, as depicted in Fig.~\ref{fig:turntype}. The majority of the frames correspond to keep-straight, while turns are required at specific intersections. We evaluate the performance for these two scenarios separately to verify the strengths of the methods under different conditions. The evaluation results (Table~\ref{tab:previousart}) reveal that Field-Beier performs better in straight-line scenarios, while Field-RRT* excels during turns. Since the metrics for turns reflect the ability to make correct road choices at intersections, we conclude that our method with  Field-RRT* planner is more advantageous for robot navigation as it prioritizes successful traversal through intersections.

\subsection{Results on Campus Dataset}

We created a campus dataset consisting of five scenes (scenes 1, 2, 3, 4, and 5) for both offline and online testing. The dataset includes LiDAR scans captured at $10$ Hz using a Livox Mid-360 LiDAR. Compared to the Velodyne HDL-64, this LiDAR produces significantly lower point density with shorter sensor range, making trajectory planning more challenging. The ground truth poses are estimated by the SLAM algorithms \cite{xu2021fast,xu2022fast,gtsam,kim2018scan}. To accommodate the new dataset, the step size, neighbor radius, and number of iterations of Field-RRT* are set to 0.75 m, 1.5 m, and 1000, respectively. The RRT tree is down-sampled half along the x- and y-axis for fast running.

\begin{table*}[htb]
  \caption{Trajectory planning comparisons of the segmentation baseline (\textbf{Se.}), end-to-end baseline (\textbf{En.}), our method with Field-RRT*(\textbf{RRT.}) and Field-Bezier (\textbf{Bez.}) on our campus dataset. All these metrics are evaluated from the current position to $10$ meters and $15$ meters ahead.}
  \label{tab:baselinesseqs2}
  \centering
  \renewcommand{\arraystretch}{1.0}
  \setlength{\tabcolsep}{4pt}
  \begin{tabular}{@{}r|cccc|cccc|cccc|cccc@{}}
  \toprule
\multicolumn{1}{c}{\multirow{2}{*}{Seq}} & \multicolumn{1}{c}{\textbf{Se.}\cite{gao2024active}} & \multicolumn{1}{c}{\textbf{En.}\cite{xu2022trajectory}} & \multicolumn{1}{c}{\textbf{RRT.}} & \multicolumn{1}{c}{\textbf{Bez.}} & \multicolumn{1}{c}{\textbf{Se.}\cite{gao2024active}} & \multicolumn{1}{c}{\textbf{En.}\cite{xu2022trajectory}} & \multicolumn{1}{c}{\textbf{RRT.}} & \multicolumn{1}{c}{\textbf{Bez.}} & \multicolumn{1}{c}{\textbf{Se.}\cite{gao2024active}} & \multicolumn{1}{c}{\textbf{En.}\cite{xu2022trajectory}} & \multicolumn{1}{c}{\textbf{RRT.}} & \multicolumn{1}{c}{\textbf{Bez.}} & \multicolumn{1}{c}{\textbf{Se.}\cite{gao2024active}} & \multicolumn{1}{c}{\textbf{En.}\cite{xu2022trajectory}} & \multicolumn{1}{c}{\textbf{RRT.}} & \multicolumn{1}{c}{\textbf{Bez.}}\\
\cmidrule(r){2-5}\cmidrule(r){6-9}\cmidrule(r){10-13}\cmidrule(r){14-17}
\cmidrule(r){2-5}\cmidrule(r){6-9}\cmidrule(r){10-13}
\multicolumn{1}{c}{} & \multicolumn{4}{c}{ADE$_{10m}\downarrow$} & \multicolumn{4}{c}{FDE$_{10m}\downarrow$} & \multicolumn{4}{c}{HitRate$_{10m}\uparrow$} & \multicolumn{4}{c}{Coverage$_{10m}\uparrow$}\\
\cmidrule(r){2-5}\cmidrule(r){6-9}\cmidrule(r){10-13}\cmidrule(r){14-17}
scene 3 & 2.52 & 0.69 & \textbf{0.56} & 0.58 & 3.98 & 1.42 & \textbf{0.90} & 0.95 & 0.27 & 0.77 & \textbf{0.91} & 0.83 & 0.50 & 0.90 & \textbf{0.96} & 0.91\\
scene 4 & 1.91 & 1.15 & 1.25 & \textbf{0.94} & 2.87 & 2.33 & 1.91 & \textbf{1.65} & 0.40 & 0.63 & 0.64 & \textbf{0.68} & 0.61 & 0.79 & 0.80 & \textbf{0.83}\\
scene 5 & 2.18 & 0.91 & 0.78 & \textbf{0.63} & 3.39 & 1.93 & 1.19 & \textbf{1.05} & 0.31 & 0.67 & \textbf{0.79} & 0.77 & 0.57 & 0.83 & \textbf{0.91} & 0.90\\
\midrule
\multicolumn{1}{c}{} & \multicolumn{4}{c}{ADE$_{15m}\downarrow$} & \multicolumn{4}{c}{FDE$_{15m}\downarrow$} & \multicolumn{4}{c}{HitRate$_{15m}\uparrow$} & \multicolumn{4}{c}{Coverage$_{15m}\uparrow$}\\
\cmidrule(r){2-5}\cmidrule(r){6-9}\cmidrule(r){10-13}\cmidrule(r){14-17}
scene 3 & 3.11 & 1.06 & \textbf{0.71} & 0.74 & 5.04 & 2.51 & \textbf{1.33} & 1.45 & 0.19 & 0.65 & \textbf{0.80} & 0.73 & 0.43 & 0.84 & \textbf{0.93} & 0.89\\
scene 4 & 2.32 & 1.77 & 1.52 & \textbf{1.25} & 3.77 & 3.82 & 2.39 & \textbf{2.26} & 0.31 & 0.51 & 0.50 & \textbf{0.56} & 0.56 & 0.71 & 0.73 & \textbf{0.77}\\
scene 5 & 2.86 & 1.45 & 0.99 & \textbf{0.88} & 5.15 & 3.13 & \textbf{1.77} & 1.86 & 0.20 & 0.53 & 0.64 & \textbf{0.65} & 0.49 & 0.74 & 0.84 & \textbf{0.85}\\
\midrule
  \end{tabular}
\end{table*}

\textbf{Offline Testing.} The experiment settings of offline testing are similar to the previous experiment on SemanticKITTI. Scenes 1 and 2 are used for training while the rest scenes are used for testing.
To train our model, no OSM routes or manual driving trajectories are provided for supervision. The free space for training is estimated by a random forest classifier trained by LiDAR points from $30$ frames. To train the end-to-end baseline model, we obtained OSM routes from the OSM website as input and ground-truth driving trajectories collected by manual driving as supervision. The segmentation baseline uses the free space estimated by the same method used for generating training samples for our model.
For evaluation, the OSM routes are downloaded from the OSM website and registered with the ground-truth driving trajectories.
During inference, these OSM routes are converted into initial OrField (Section \ref{sec:ngm}) for network input. 
The results are shown in Table \ref{tab:baselinesseqs2} from which we see that our method achieves supreme performance again across all the testing scenes for all the testing metrics. This shows that our method is also robust to low-density data and short sensor range.

\begin{figure*}[htb!]
    \centering
    \includegraphics[width=1.0\linewidth]{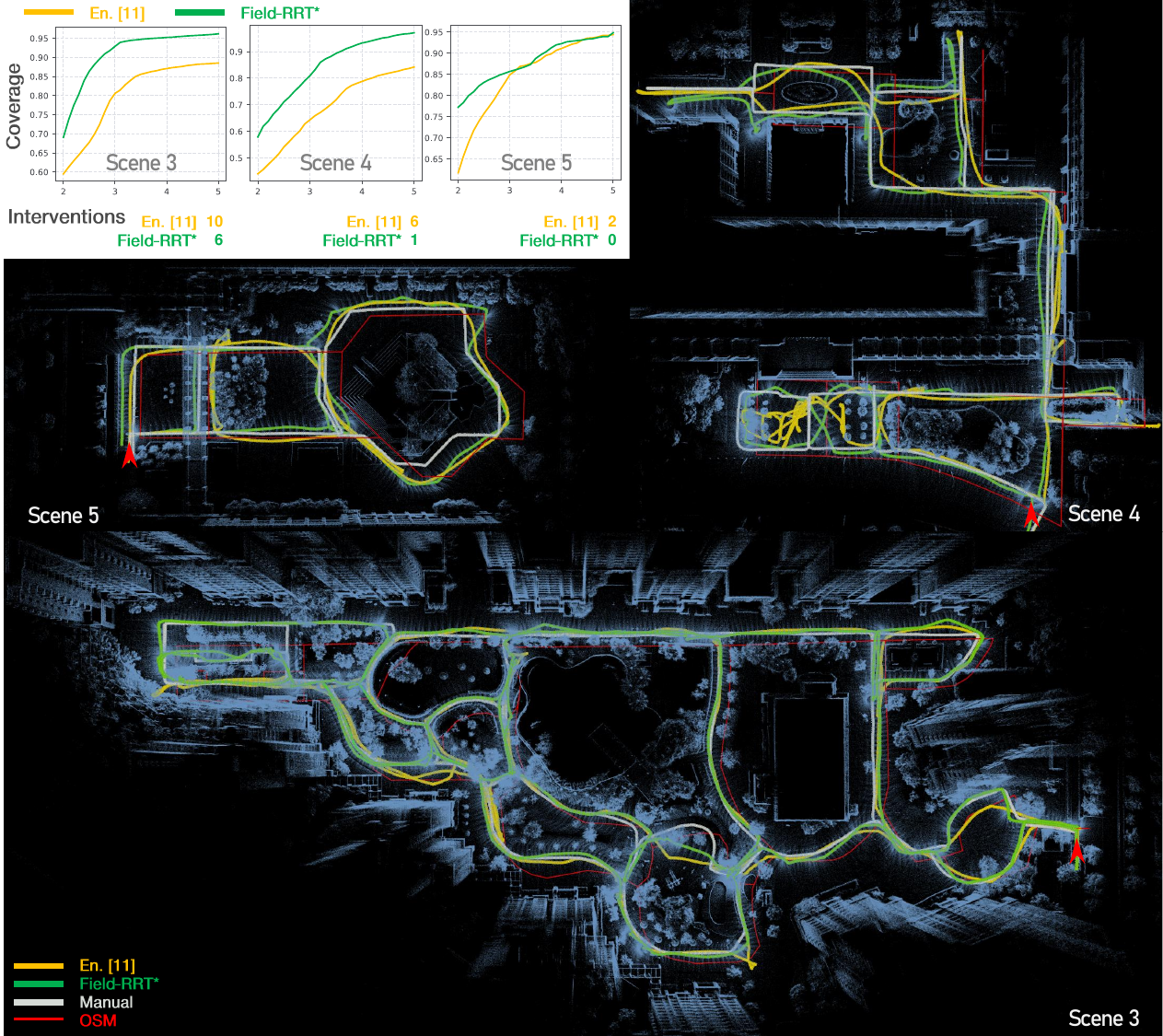}
    \caption{\textbf{Top Left:} The ratio of the area covered by the automatic vehicle to the area covered by OSM routes. During the traversal, we intervene when the robot enters an infeasible branch or gets stuck while moving forward. \textbf{Top Right and Bottom:} The visualization of automatic traversals with OSM routes on our campus scenes. 
        The manually control trajectories are shown in white. The OSM routes are shown in red. The start poses are shown in the red arrow.}
    \label{fig:realrobot}
\end{figure*}

\textbf{Online Testing.} To evaluate the methods in real-world applications, we conduct online experiments where trained models are tested on live data in scenes 3, 4, and 5, which cover large, medium, and small areas, respectively, as shown in Fig. \ref{fig:realrobot}. The experiments are conducted on campus using a ground vehicle equipped with an Intel NUC 11 computer. 
To assess online performance, we calculate the \textbf{Coverage} after the vehicle automatically traverses a scene and returns to the starting position. Specifically, we at first manually control the vehicle to perfectly traverse the scene and record the driving trajectory. By removing the trajectory points that duplicate in their nearby, we get a collection that contains the trajectory points defining the area covered by OSM routes. For each point in the collection, we define it as covered if the traversal trajectory of the automatic vehicle intersects the circle centered at the point with a radius of $r$ meters. Then we calculate the Coverage as the ratio of the number of the covered points to the number of all the points in the collection. 
Besides, we introduce a metric called \textbf{Intervention}, defined as the number of manual interventions required during the vehicle traversal. An intervention occurs when the vehicle fails during navigation and must be manually steered onto the correct track. Failures include getting stuck in road ditches or choosing infeasible branches at intersections when following the OSM trajectory, preventing successful navigation.
The experiments were carried out three times for each scene at three different starting positions. For calculating Coverage, we set $r$ from $2$ meters to $5$ meters every $0.1$ meter and plot the result for each $r$. The results of average Coverage and Interventions are shown in Fig.~\ref{fig:realrobot}.
Our method demonstrates fewer interventions and achieves a larger coverage of the target area compared to the end-to-end baseline \cite{xu2022trajectory}. This is mainly because our approach excels in selecting the correct branches at intersections, even when the OSM routes significantly deviate from actual roads. Additionally, our method can execute turnarounds after encountering dead ends while following OSM routes. In contrast, the baseline struggles with turnarounds due to the absence of such behavior in its supervision. Furthermore, the baseline tends to predict smooth trajectories extrapolated from previous vehicle poses, which are often inconsistent with the precise trajectories required in narrow lanes. Figure \ref{fig:realrobot} visualizes the trajectories and maps of the evaluated scenes, showcasing the advantages of our approach in real-world scenarios.

\subsection{Ablation Studies for Orientations}
\label{sec:ablations}

In this section, we do ablation studies to confirm the best one among different OrFields including those derived before and after optimization by our deep network.  
Meanwhile, we investigate the superiority of planning with orientation representation over the binary free space representation without the orientation information.

\textbf{Before Network Optimization}. 
The OrFields before optimization by our deep network are categorized into two types based on whether they incorporate sensor information.
For the OrFields incorporating sensor information, we estimate the free space using environmental data collected from onboard sensors. Based on this free space estimation, orientations are calculated using the Dijkstra algorithm from the local target point, referred to as \textbf{Dijkstra}. Section~\ref{sec:optimizedfiled} introduces how orientations for training samples are generated by operating on the gradient of the free space. Using the same method, orientations are calculated here and denoted as \textbf{Gradient}. To highlight the importance of future frames, we aggregate them with the past and current frames to estimate free space. Orientations are then generated using the Dijkstra and Gradient methods, denoted as \textbf{Dijkstra-Full} and \textbf{Gradient-Full}, respectively. These OrFields, which leverage future frames, are unavailable during real-world deployment. Thus, they serve as a reference upper bound for the performance of OrFields without deep network optimization.

Another approach is to estimate orientations solely based on OSM routes, without incorporating sensor information. For this, the orientation of a grid is assigned as the tangent of its nearest OSM route edge, calculated from the OSM topology.  We denote this method as \textbf{Nearest}.
Additionally, the initial OrField for network input in Section~\ref{sec:ngm}, referred to as \textbf{InitialOrField}, is a smoothed version of orientations derived from the OSM topology.

Table~\ref{tab:ablations} highlights several findings: 1. The improvement of Gradient (Gr.) over Dijkstra (Di.) demonstrates the validity of calculating orientations from gradient directions, as discussed in Section~\ref{sec:optimizedfiled}.
2. Aggregating future frames reduces the loss of surrounding information. The improvement from Gradient (Gr.) to Gradient-Full (Gr.$^\prime$) illustrates this point and emphasizes the importance of full observations in trajectory planning. This further supports the rationale for designing a network to estimate orientations from partial observations. 3. Meanwhile, the underperformance of Nearest (Ne.) and InitialOrField (IOr.) shows the necessity of optimization on the orientations derived solely from OSM routes with the existence of noise.

\begin{table}[tb!]
  \caption{Error analysis between planned trajectory and manual driving trajectory on SemanticKITTI dataset.
  \textbf{Mo.} denotes {Modular}. 
  \textbf{Di.} denotes {Dijkstra}. \textbf{Gr.} denotes {Gradient}. \textbf{Di}.$^\prime$ denotes{Dijkstra-Full}. \textbf{Gr.}$^\prime$ denotes {Gradient-Full}. \textbf{Ne.} denotes {Nearest}. \textbf{IOr.} denotes {InitialOrField}. \textbf{FS-DOr.} denotes {FS-DeepOrField}. \textbf{DOr.} denotes {DeepOrField}. \textbf{RRT.} denotes {Field-RRT*}. \textbf{Bez.} denotes {Field-Beizer}. 
  All metrics are evaluated from the current position to $10$ meters and $20$ meters ahead.}
  \label{tab:ablations}
  \centering
  \renewcommand{\arraystretch}{1.0}
  \setlength{\tabcolsep}{3pt}
  \resizebox{0.98\columnwidth}{!}{
  \begin{tabular}{@{}r|cccccccccc@{}}
\toprule
\multicolumn{1}{c}{} & \multicolumn{1}{c}{\textbf{Point}} & \multicolumn{9}{c}{\textbf{Field}}\\
\cmidrule(r){2-2}\cmidrule(r){3-11}
\multicolumn{1}{c}{} & \multicolumn{1}{c}{} & \multicolumn{6}{c}{\textbf{Before Opt.}} & \multicolumn{3}{c}{\textbf{After Opt.}}\\
\cmidrule(r){3-8}\cmidrule(r){9-11}
\multicolumn{1}{c}{} & \multicolumn{1}{c}{\textbf{RRT*}} & \multicolumn{8}{c}{\textbf{RRT.}} & \multicolumn{1}{c}{\textbf{Bez.}}\\
\cmidrule(r){2-2}\cmidrule(r){3-10}\cmidrule(r){11-11}
\multicolumn{1}{c}{\multirow{2}{*}{Seq}} & \multicolumn{1}{c}{\textbf{Mo.}} & \multicolumn{1}{c}{\textbf{Di.}} & \multicolumn{1}{c}{\textbf{Di.}$^\prime$} & \multicolumn{1}{c}{\textbf{Gr.}} & \multicolumn{1}{c}{\textbf{Gr.}$^\prime$} & \multicolumn{1}{c}{\textbf{Ne.}} & \multicolumn{1}{c}{\textbf{IOr.}} & \multicolumn{1}{c}{\textbf{FS-DOr.}} & \multicolumn{1}{c}{\textbf{DOr.}} & \multicolumn{1}{c}{\textbf{DOr.}}\\
\cmidrule(r){2-11}
\multicolumn{1}{c}{} & \multicolumn{10}{c}{ADE$_{10m}\downarrow$}\\
\cmidrule(r){2-11}
08 & 0.53 & 0.89 & 0.95 & 0.65 & 0.52 & 0.61 & 0.62 & 0.26 & 0.25 & \textbf{0.21}\\
13 & 0.93 & 1.35 & 1.15 & 0.97 & 0.61 & 0.57 & 0.79 & 0.26 & 0.27 & \textbf{0.24}\\
15 & 0.42 & 0.71 & 0.77 & 0.56 & 0.51 & 0.61 & 0.81 & 0.30 & 0.26 & \textbf{0.22}\\
16 & 0.50 & 1.05 & 1.11 & 0.67 & 0.34 & 0.48 & 0.55 & 0.21 & 0.21 & \textbf{0.19}\\
18 & 0.41 & 0.65 & 0.72 & 0.67 & 0.56 & 0.59 & 0.32 & 0.19 & 0.16 & \textbf{0.13}\\
19 & 0.53 & 1.13 & 1.14 & 0.69 & 0.61 & 0.90 & 0.72 & 0.31 & 0.27 & \textbf{0.21}\\
\midrule
\multicolumn{1}{c}{} & \multicolumn{10}{c}{ADE$_{20m}\downarrow$}\\
\cmidrule(r){2-11}
08 & 0.98 & 1.11 & 1.20 & 0.86 & 0.79 & 1.10 & 1.19 & 0.49 & 0.44 & \textbf{0.41}\\
13 & 1.69 & 1.89 & 1.47 & 1.52 & 0.89 & 1.07 & 1.58 & 0.50 & 0.44 & \textbf{0.43}\\
15 & 0.74 & 0.84 & 0.96 & 0.76 & 0.84 & 1.12 & 1.60 & 0.57 & 0.44 & \textbf{0.37}\\
16 & 0.96 & 1.23 & 1.33 & 0.76 & 0.49 & 0.91 & 1.11 & 0.40 & 0.37 & \textbf{0.36}\\
18 & 0.81 & 0.88 & 0.81 & 0.98 & 0.78 & 1.13 & 0.71 & 0.46 & 0.38 & \textbf{0.30}\\
19 & 1.10 & 1.35 & 1.39 & 0.87 & 0.81 & 1.73 & 1.40 & 0.53 & 0.45 & \textbf{0.42}\\
\midrule
  \end{tabular}
  }
\end{table}

\begin{figure*}[ht!]
    \centering
    \includegraphics[width=1.0\linewidth]{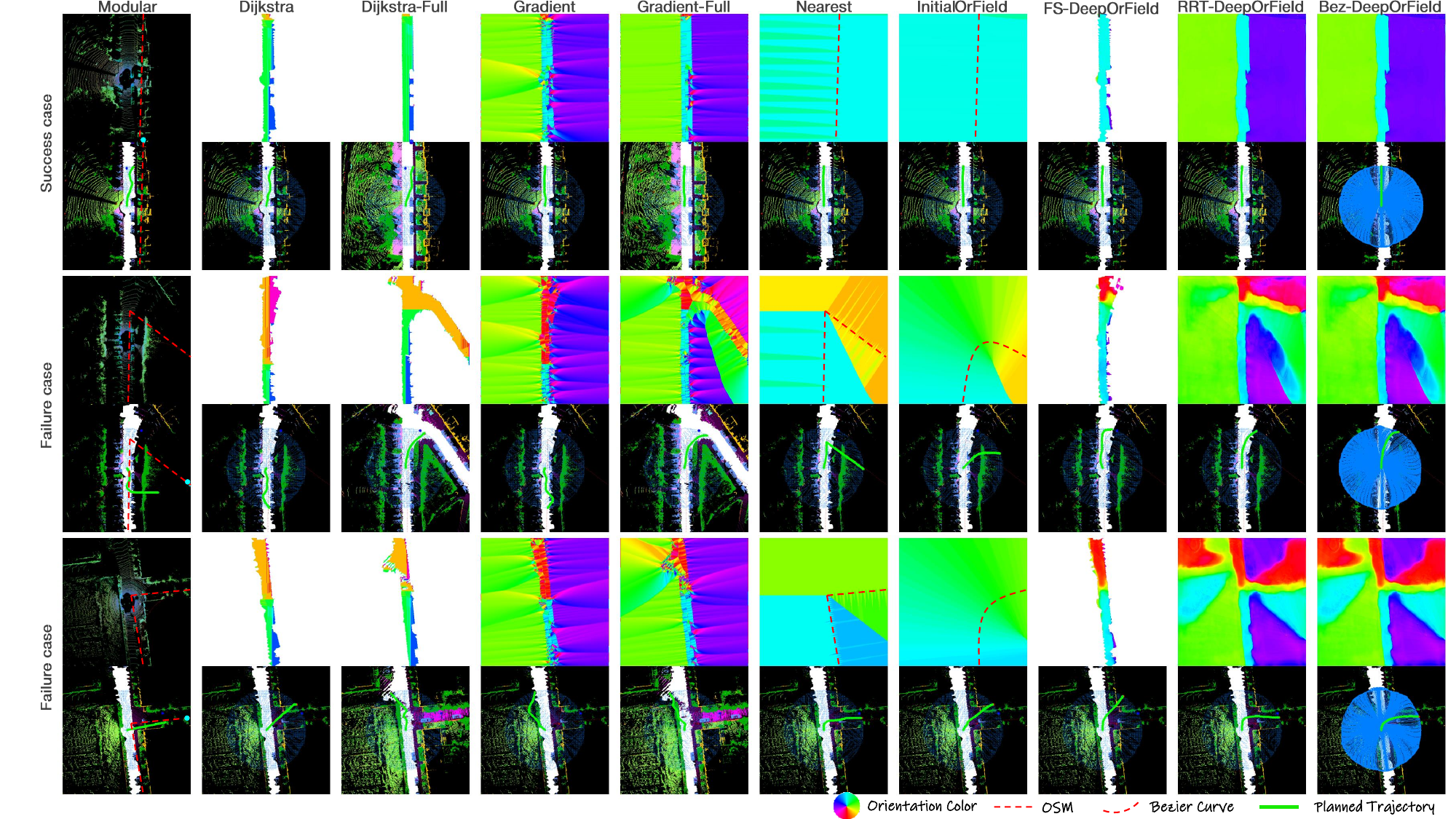}
    \caption{Examples of trajectories estimated with different orientations on SemanticKITTI dataset. From left to right are \textbf{Modular}, \textbf{Dijkstra}, \textbf{Dijkstra-Full}, \textbf{Gradient}, \textbf{Gradient-Full}, \textbf{Nearest}, \textbf{InitialOrField}, \textbf{FS-DeepOrField}, \textbf{DeepOrField} with Field-RRT*, and \textbf{DeepOrField} with Field-Bezier, respectively. The top row shows success cases for all comparing methods. The bottom two rows show failure cases for methods based on free space estimation while our methods still achieve good results.}
    \label{fig:cases}
\end{figure*}

\textbf{OrField After Network Optimization.} 
Following the evaluation of heuristically designed OrFields without network optimization, we analyze the performance of OrFields after optimization by our deep network, referred to as \textbf{DeepOrField}. 
If free space information is available, it can be integrated into DeepOrField by setting the orientation length in non-free space areas to zero, thereby discouraging traversal in those regions. This variant is denoted as \textbf{FS-DeepOrField}. 

Table~\ref{tab:ablations} shows a clear performance gap between the orientations before and after optimization, demonstrating the effectiveness of our deep network. When comparing the optimized orientations, the integration of occupancy information in FS-DeepOrField (FS-DOr.) resulted in a degraded performance in most sequences compared to using all orientations from the network in DeepOrField (DOr.). This indicates our deep network provides valuable orientation information not only in the free space but also in the obstacle and unobserved area. Our optimized orientations are more accurate than the "oracle" one Gradient-Full (Gr.$^\prime$) which sees the future frames. 

\textbf{Point Planner} versus \textbf{Field Planner}. The point planner is a traditional RRT* planner that approaches a target point by expanding an exploration tree following the estimation of the free space and the selection of the local target in a step-by-step manner akin to~\cite{suger2017global, ort2018autonomous}, denoted as \textbf{Modular}. Instead, our field planner expands the tree constrained by the orientations along the edge. We demonstrate the superiority of our field planner over the point planner by comparison.

The results are shown in Table~\ref{tab:ablations} from which we see that compared to the point planner (Mo.), our field planner (Gr.) which incorporates constraints on orientation achieves better performance when evaluated at distances of $20$ meters ahead. 
This aligns with the findings of previous studies~\cite{ort2019maplite, omama2023alt}, which optimized trajectories using the Euclidean Distance Transform (EDT) to encourage paths that align with the road centerline. 
Meanwhile, this indicates that the field planner outperforms the traditional point planner in scenarios requiring planning over longer distances which are more susceptible to occlusion, encouraging us to design a network to estimate more accurate orientations both within and beyond the sensor range.

Based on these ablation studies, we conclude that our method significantly enhances orientation accuracy, even under partial observations and noisy OSM data. Examples of both successful and failure cases are illustrated in Fig. \ref{fig:cases}.

\begin{table}[htb]
\caption{Error analysis between planned trajectory and manual driving trajectory on SemanticKITTI dataset.
  \textbf{Full} denotes using the LiDAR BEV, initial OrField, and distance map for network input. \textbf{-Distance} denotes not using the distance map. \textbf{-Init} denotes not using the initial OrField. \textbf{RRT.} denotes {Field-RRT*}. \textbf{Bez.} denotes {Field-Beizer}. 
  All metrics are evaluated from the current position to $10$ meters and $20$ meters ahead.}
\label{tab:ablations2}
\centering
\renewcommand{\arraystretch}{1.0}
\setlength{\tabcolsep}{8pt}
\begin{tabular}{@{}rccc|ccc@{}}
\toprule
\multicolumn{1}{c}{} & \multicolumn{1}{c}{\textbf{Full}} & \multicolumn{1}{c}{-\textbf{Distance}} & \multicolumn{1}{c}{-\textbf{Init}} & \multicolumn{1}{c}{\textbf{Full}} & \multicolumn{1}{c}{-\textbf{Distance}} & \multicolumn{1}{c}{-\textbf{Init}}\\
\cmidrule(r){2-4}\cmidrule(r){5-7}
\multicolumn{1}{c}{\multirow{2}{*}{Seq}} & \multicolumn{3}{c}{\textbf{RRT.}} & \multicolumn{3}{c}{\textbf{Bez.}}\\
\cmidrule(r){2-4}\cmidrule(r){5-7}
\multicolumn{1}{c}{} & \multicolumn{6}{c}{ADE$_{10m}\downarrow$}\\
\cmidrule(r){2-7}
08 & \textbf{0.25} & 0.26 & 0.32 & \textbf{0.21} & \textbf{0.21} & 0.26\\
13 & \textbf{0.27} & \textbf{0.27} & \textbf{0.27} & \textbf{0.24} & \textbf{0.24} & \textbf{0.24}\\
15 & 0.26 & \textbf{0.24} & 0.29 & \textbf{0.22} & \textbf{0.22} & 0.27\\
16 & 0.21 & \textbf{0.19} & 0.24 & \textbf{0.19} & \textbf{0.19} & 0.20\\
18 & \textbf{0.16} & 0.18 & 0.32 & \textbf{0.13} & 0.18 & 0.32\\
19 & \textbf{0.27} & 0.31 & 0.38 & \textbf{0.21} & 0.22 & 0.26\\
\midrule
\multicolumn{1}{c}{} & \multicolumn{6}{c}{ADE$_{20m}\downarrow$}\\
\cmidrule(r){2-7}
08 & \textbf{0.44} & 0.47 & 0.67 & \textbf{0.41} & 0.43 & 0.62\\
13 & \textbf{0.44} & 0.46 & 0.52 & \textbf{0.43} & 0.46 & 0.49\\
15 & 0.44 & \textbf{0.42} & 0.67 & 0.37 & \textbf{0.36} & 0.61\\
16 & 0.37 & \textbf{0.35} & 0.47 & \textbf{0.36} & 0.37 & 0.41\\
18 & \textbf{0.38} & 0.44 & 0.82 & \textbf{0.30} & 0.38 & 0.79\\
19 & \textbf{0.45} & 0.48 & 0.68 & \textbf{0.42} & 0.44 & 0.59\\
\midrule
\end{tabular}
\end{table}

\subsection{Ablation Studies for Network Inputs}
\label{sec:ablations2}

The inputs for the network consist of the LiDAR BEV, initial OrField, and distance map. To investigate the best chosen for the network inputs, we do ablation studies by removing the distance map and initial OrField, respectively. The network loses the position information about the OSM route without the distance map, while the network loses the orientation information about the OSM route without the initial OrField. The ablation results are shown in Table \ref{tab:ablations2} from which we can see a clear improvement when using the initial OrField as network input. Although the distance map is enough to prompt the underlying direction for navigation, it is difficult to decide when the OSM route renders large uncertainties. Besides, the improvement when using the distance map, especially in sequences 08, 13, and 19 of large scale, indicates the usefulness of incorporating it because provides the confidence of position of the OSM route.

\begin{figure*}[htb!]
    \centering
    \includegraphics[width=1.0\linewidth]{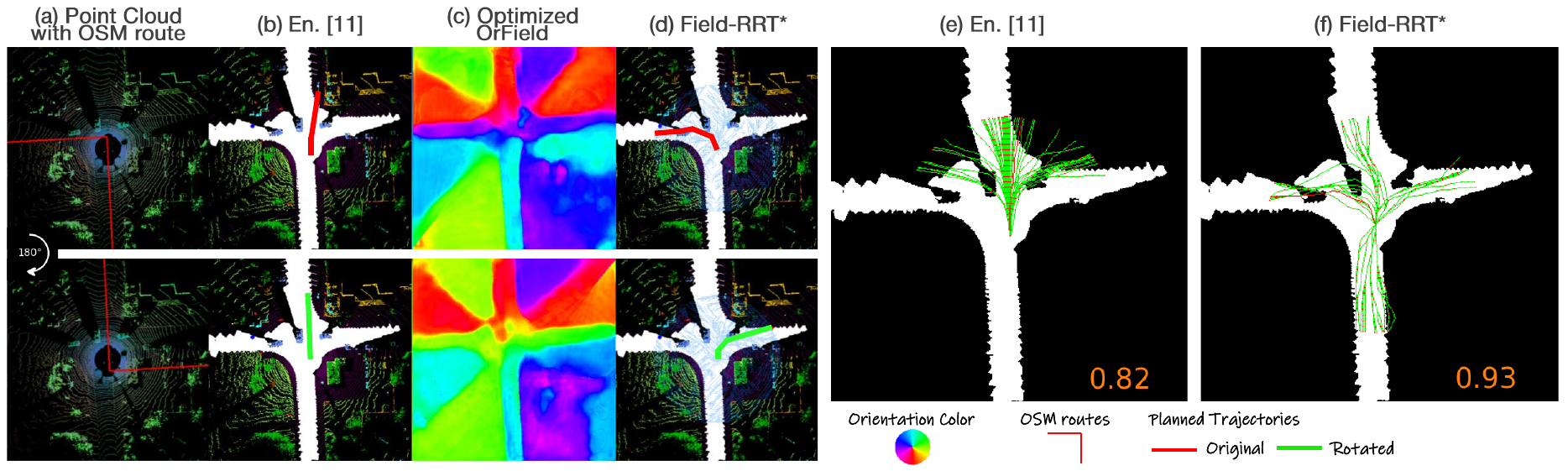}
    \caption{The comparison of our method with the end-to-end baseline \cite{xu2022trajectory} in a 4-branches intersection on SemanticKITTI sequence 08. The OSM route is rotated with different angles. The first row in (a,b,c,d): The cases rotated with $0$ degree. The second row in (a,b,c,d): The cases rotated with $180$ degree. (e,f): The planned path for all the cases.}
    \label{fig:robustness}
\end{figure*}

\subsection{Robustness Analysis}
We analyze the robustness of our method that plans with OrField compared with the end-to-end trajectory planning method in different cases at the same intersection. Specifically, we generate different cases by rotating the OSM route from $0$ to $360$ degree every $6$ degree at a typical 4-branches intersection. In this way, $60$ cases are generated, in which OSM routes lead to different branches. An example of an original OSM route and the one rotated by $180$ degree are shown in the first and second rows of Fig. \ref{fig:robustness} (a), respectively. The paths planned by the end-to-end baseline \cite{xu2022trajectory} method are shown in Fig. \ref{fig:robustness} (b). We can see that the baseline predicts forward paths, while the OSM route indicates a left turn and a right turn. Instead, our method plans the correct path, as shown in Fig. \ref{fig:robustness} (d). 
For evaluation, we collect all the planned paths in all the cases with different rotated angles and plot them in Fig. \ref{fig:robustness} (e) and (f). We find the paths most forward for the baseline and are separated into different branches for our method. This means that the distribution of paths planned by our method goes beyond the biased distribution learned from manual driving. We calculate the percentage of the points of planned paths lying in the free space. As shown in Fig. \ref{fig:robustness} (e) and (f), the baseline plans many paths across the obstacles and only $82\%$ planned points lying in free space, while our method plans most paths lying in the free space and achieves $93\%$. We find similar conclusions in other intersections.

\section{Conclusion}
In this paper, we propose a novel representation for navigation, named Orientation Field (OrField). We demonstrate the superiority of OrField while using it as a constraint for trajectory planning. To optimize the OrField jointly from LiDAR scans and OSM route, we propose a deep learning network to optimize the initial OrField solely derived from OSM route. Combined with the optimized OrField, Field Planners are proposed to find the best path. Experiments show the superiority of our optimized OrField compared with step-by-step ways to generate trajectory and the previous end-to-end work that predicts the waypoints. Ablations show the effectiveness of our deep network to optimize the OrField compared with generating OrField in rule-based ways. 

\bibliographystyle{IEEEtran}
\bibliography{main}

\begin{thebibliography}{10}
\providecommand{\url}[1]{#1}
\csname url@samestyle\endcsname
\providecommand{\newblock}{\relax}
\providecommand{\bibinfo}[2]{#2}
\providecommand{\BIBentrySTDinterwordspacing}{\spaceskip=0pt\relax}
\providecommand{\BIBentryALTinterwordstretchfactor}{4}
\providecommand{\BIBentryALTinterwordspacing}{\spaceskip=\fontdimen2\font plus
\BIBentryALTinterwordstretchfactor\fontdimen3\font minus \fontdimen4\font\relax}
\providecommand{\BIBforeignlanguage}[2]{{%
\expandafter\ifx\csname l@#1\endcsname\relax
\typeout{** WARNING: IEEEtran.bst: No hyphenation pattern has been}%
\typeout{** loaded for the language `#1'. Using the pattern for}%
\typeout{** the default language instead.}%
\else
\language=\csname l@#1\endcsname
\fi
#2}}
\providecommand{\BIBdecl}{\relax}
\BIBdecl

\bibitem{bao2023review}
Z.~Bao, S.~Hossain, H.~Lang, and X.~Lin, ``A review of high-definition map creation methods for autonomous driving,'' \emph{Engineering Applications of Artificial Intelligence}, vol. 122, p. 106125, 2023.

\bibitem{zeng2019end}
W.~Zeng, W.~Luo, S.~Suo, A.~Sadat, B.~Yang, S.~Casas, and R.~Urtasun, ``End-to-end interpretable neural motion planner,'' in \emph{Proceedings of the IEEE/CVF Conference on Computer Vision and Pattern Recognition}, 2019, pp. 8660--8669.

\bibitem{sadat2020perceive}
A.~Sadat, S.~Casas, M.~Ren, X.~Wu, P.~Dhawan, and R.~Urtasun, ``Perceive, predict, and plan: Safe motion planning through interpretable semantic representations,'' in \emph{Computer Vision--ECCV 2020: 16th European Conference, Glasgow, UK, August 23--28, 2020, Proceedings, Part XXIII 16}.\hskip 1em plus 0.5em minus 0.4em\relax Springer, 2020, pp. 414--430.

\bibitem{haklay2008openstreetmap}
M.~Haklay and P.~Weber, ``Openstreetmap: User-generated street maps,'' \emph{IEEE Pervasive computing}, vol.~7, no.~4, pp. 12--18, 2008.

\bibitem{hentschel2010autonomous}
M.~Hentschel and B.~Wagner, ``Autonomous robot navigation based on openstreetmap geodata,'' in \emph{13th International IEEE Conference on Intelligent Transportation Systems}.\hskip 1em plus 0.5em minus 0.4em\relax IEEE, 2010, pp. 1645--1650.

\bibitem{floros2013openstreetslam}
G.~Floros, B.~Van Der~Zander, and B.~Leibe, ``Openstreetslam: Global vehicle localization using openstreetmaps,'' in \emph{2013 IEEE international conference on robotics and automation}.\hskip 1em plus 0.5em minus 0.4em\relax IEEE, 2013, pp. 1054--1059.

\bibitem{suger2017global}
B.~Suger and W.~Burgard, ``Global outer-urban navigation with openstreetmap,'' in \emph{2017 IEEE International Conference on Robotics and Automation (ICRA)}.\hskip 1em plus 0.5em minus 0.4em\relax IEEE, 2017, pp. 1417--1422.

\bibitem{ort2018autonomous}
T.~Ort, L.~Paull, and D.~Rus, ``Autonomous vehicle navigation in rural environments without detailed prior maps,'' in \emph{2018 IEEE international conference on robotics and automation (ICRA)}.\hskip 1em plus 0.5em minus 0.4em\relax IEEE, 2018, pp. 2040--2047.

\bibitem{ort2019maplite}
T.~Ort, K.~Murthy, R.~Banerjee, S.~K. Gottipati, D.~Bhatt, I.~Gilitschenski, L.~Paull, and D.~Rus, ``Maplite: Autonomous intersection navigation without a detailed prior map,'' \emph{IEEE Robotics and Automation Letters}, vol.~5, no.~2, pp. 556--563, 2019.

\bibitem{omama2023alt}
M.~Omama, P.~Inani, P.~Paul, S.~C. Yellapragada, K.~M. Jatavallabhula, S.~Chinchali, and M.~Krishna, ``Alt-pilot: Autonomous navigation with language augmented topometric maps,'' \emph{arXiv preprint arXiv:2310.02324}, 2023.

\bibitem{xu2022trajectory}
J.~Xu, L.~Xiao, D.~Zhao, Y.~Nie, and B.~Dai, ``Trajectory prediction for autonomous driving with topometric map,'' in \emph{2022 International Conference on Robotics and Automation (ICRA)}.\hskip 1em plus 0.5em minus 0.4em\relax IEEE, 2022, pp. 8403--8408.

\bibitem{paz2021tridentnet}
D.~Paz, H.~Zhang, and H.~I. Christensen, ``Tridentnet: A conditional generative model for dynamic trajectory generation,'' in \emph{International Conference on Intelligent Autonomous Systems}.\hskip 1em plus 0.5em minus 0.4em\relax Springer, 2021, pp. 403--416.

\bibitem{paz2022tridentnetv2}
D.~Paz, H.~Xiang, A.~Liang, and H.~I. Christensen, ``Tridentnetv2: Lightweight graphical global plan representations for dynamic trajectory generation,'' in \emph{2022 International Conference on Robotics and Automation (ICRA)}.\hskip 1em plus 0.5em minus 0.4em\relax IEEE, 2022, pp. 9265--9271.

\bibitem{guo2022end}
K.~Guo, W.~Liu, and J.~Pan, ``End-to-end trajectory distribution prediction based on occupancy grid maps,'' in \emph{Proceedings of the IEEE/CVF Conference on Computer Vision and Pattern Recognition}, 2022, pp. 2242--2251.

\bibitem{cui2019multimodal}
H.~Cui, V.~Radosavljevic, F.-C. Chou, T.-H. Lin, T.~Nguyen, T.-K. Huang, J.~Schneider, and N.~Djuric, ``Multimodal trajectory predictions for autonomous driving using deep convolutional networks,'' in \emph{2019 International Conference on Robotics and Automation (ICRA)}.\hskip 1em plus 0.5em minus 0.4em\relax IEEE, 2019, pp. 2090--2096.

\bibitem{chai2019multipath}
Y.~Chai, B.~Sapp, M.~Bansal, and D.~Anguelov, ``Multipath: Multiple probabilistic anchor trajectory hypotheses for behavior prediction,'' \emph{arXiv preprint arXiv:1910.05449}, 2019.

\bibitem{phan2020covernet}
T.~Phan-Minh, E.~C. Grigore, F.~A. Boulton, O.~Beijbom, and E.~M. Wolff, ``Covernet: Multimodal behavior prediction using trajectory sets,'' in \emph{Proceedings of the IEEE/CVF conference on computer vision and pattern recognition}, 2020, pp. 14\,074--14\,083.

\bibitem{li2021openstreetmap}
J.~Li, H.~Qin, J.~Wang, and J.~Li, ``Openstreetmap-based autonomous navigation for the four wheel-legged robot via 3d-lidar and ccd camera,'' \emph{IEEE Transactions on Industrial Electronics}, vol.~69, no.~3, pp. 2708--2717, 2021.

\bibitem{bolles1981ransac}
R.~C. Bolles and M.~A. Fischler, ``A ransac-based approach to model fitting and its application to finding cylinders in range data.'' in \emph{IJCAI}, vol. 1981, 1981, pp. 637--643.

\bibitem{dosovitskiy2017carla}
A.~Dosovitskiy, G.~Ros, F.~Codevilla, A.~Lopez, and V.~Koltun, ``Carla: An open urban driving simulator,'' in \emph{Conference on robot learning}.\hskip 1em plus 0.5em minus 0.4em\relax PMLR, 2017, pp. 1--16.

\bibitem{werling2010optimal}
M.~Werling, J.~Ziegler, S.~Kammel, and S.~Thrun, ``Optimal trajectory generation for dynamic street scenarios in a frenet frame,'' in \emph{2010 IEEE international conference on robotics and automation}.\hskip 1em plus 0.5em minus 0.4em\relax IEEE, 2010, pp. 987--993.

\bibitem{tsiakas2023leveraging}
K.~Tsiakas, D.~Alexiou, D.~Giakoumis, A.~Gasteratos, and D.~Tzovaras, ``Leveraging multimodal sensing and topometric mapping for human-like autonomous navigation in complex environments,'' in \emph{2023 IEEE/RSJ International Conference on Intelligent Robots and Systems (IROS)}.\hskip 1em plus 0.5em minus 0.4em\relax IEEE, 2023, pp. 7415--7421.

\bibitem{gu2023vip3d}
J.~Gu, C.~Hu, T.~Zhang, X.~Chen, Y.~Wang, Y.~Wang, and H.~Zhao, ``Vip3d: End-to-end visual trajectory prediction via 3d agent queries,'' in \emph{Proceedings of the IEEE/CVF Conference on Computer Vision and Pattern Recognition}, 2023, pp. 5496--5506.

\bibitem{zhao2021tnt}
H.~Zhao, J.~Gao, T.~Lan, C.~Sun, B.~Sapp, B.~Varadarajan, Y.~Shen, Y.~Shen, Y.~Chai, C.~Schmid \emph{et~al.}, ``Tnt: Target-driven trajectory prediction,'' in \emph{Conference on Robot Learning}.\hskip 1em plus 0.5em minus 0.4em\relax PMLR, 2021, pp. 895--904.

\bibitem{gu2021densetnt}
J.~Gu, C.~Sun, and H.~Zhao, ``Densetnt: End-to-end trajectory prediction from dense goal sets,'' in \emph{Proceedings of the IEEE/CVF International Conference on Computer Vision}, 2021, pp. 15\,303--15\,312.

\bibitem{liu2024laformer}
M.~Liu, H.~Cheng, L.~Chen, H.~Broszio, J.~Li, R.~Zhao, M.~Sester, and M.~Y. Yang, ``Laformer: Trajectory prediction for autonomous driving with lane-aware scene constraints,'' in \emph{Proceedings of the IEEE/CVF Conference on Computer Vision and Pattern Recognition}, 2024, pp. 2039--2049.

\bibitem{chitta2022transfuser}
K.~Chitta, A.~Prakash, B.~Jaeger, Z.~Yu, K.~Renz, and A.~Geiger, ``Transfuser: Imitation with transformer-based sensor fusion for autonomous driving,'' \emph{IEEE Transactions on Pattern Analysis and Machine Intelligence}, vol.~45, no.~11, pp. 12\,878--12\,895, 2022.

\bibitem{yang2023iplanner}
F.~Yang, C.~Wang, C.~Cadena, and M.~Hutter, ``iplanner: Imperative path planning,'' \emph{arXiv preprint arXiv:2302.11434}, 2023.

\bibitem{deo2020trajectory}
N.~Deo and M.~M. Trivedi, ``Trajectory forecasts in unknown environments conditioned on grid-based plans,'' \emph{arXiv preprint arXiv:2001.00735}, 2020.

\bibitem{shah2023gnm}
D.~Shah, A.~Sridhar, A.~Bhorkar, N.~Hirose, and S.~Levine, ``Gnm: A general navigation model to drive any robot,'' in \emph{2023 IEEE International Conference on Robotics and Automation (ICRA)}.\hskip 1em plus 0.5em minus 0.4em\relax IEEE, 2023, pp. 7226--7233.

\bibitem{cortinhal2020salsanext}
T.~Cortinhal, G.~Tzelepis, and E.~Erdal~Aksoy, ``Salsanext: Fast, uncertainty-aware semantic segmentation of lidar point clouds,'' in \emph{Advances in Visual Computing: 15th International Symposium, ISVC 2020, San Diego, CA, USA, October 5--7, 2020, Proceedings, Part II 15}.\hskip 1em plus 0.5em minus 0.4em\relax Springer, 2020, pp. 207--222.

\bibitem{YuKoltun2016}
F.~Yu and V.~Koltun, ``Multi-scale context aggregation by dilated convolutions,'' in \emph{ICLR}, 2016.

\bibitem{behley2019semantickitti}
J.~Behley, M.~Garbade, A.~Milioto, J.~Quenzel, S.~Behnke, C.~Stachniss, and J.~Gall, ``Semantickitti: A dataset for semantic scene understanding of lidar sequences,'' in \emph{Proceedings of the IEEE/CVF international conference on computer vision}, 2019, pp. 9297--9307.

\bibitem{gao2024active}
W.~Gao, Z.~Sun, M.~Zhao, C.-Z. Xu, and H.~Kong, ``Active loop closure for osm-guided robotic mapping in large-scale urban environments,'' in \emph{2024 IEEE/RSJ International Conference on Intelligent Robots and Systems (IROS)}.\hskip 1em plus 0.5em minus 0.4em\relax IEEE, 2024, pp. 12\,302--12\,309.

\bibitem{geiger2012we}
A.~Geiger, P.~Lenz, and R.~Urtasun, ``Are we ready for autonomous driving? the kitti vision benchmark suite,'' in \emph{2012 IEEE conference on computer vision and pattern recognition}.\hskip 1em plus 0.5em minus 0.4em\relax IEEE, 2012, pp. 3354--3361.

\bibitem{xu2021fast}
W.~Xu and F.~Zhang, ``Fast-lio: A fast, robust lidar-inertial odometry package by tightly-coupled iterated kalman filter,'' \emph{IEEE Robotics and Automation Letters}, vol.~6, no.~2, pp. 3317--3324, 2021.

\bibitem{xu2022fast}
W.~Xu, Y.~Cai, D.~He, J.~Lin, and F.~Zhang, ``Fast-lio2: Fast direct lidar-inertial odometry,'' \emph{IEEE Transactions on Robotics}, vol.~38, no.~4, pp. 2053--2073, 2022.

\bibitem{gtsam}
\BIBentryALTinterwordspacing
F.~Dellaert and G.~Contributors, ``borglab/gtsam,'' May 2022. [Online]. Available: \url{https://github.com/borglab/gtsam)}
\BIBentrySTDinterwordspacing

\bibitem{kim2018scan}
G.~Kim and A.~Kim, ``Scan context: Egocentric spatial descriptor for place recognition within 3d point cloud map,'' in \emph{2018 IEEE/RSJ International Conference on Intelligent Robots and Systems (IROS)}.\hskip 1em plus 0.5em minus 0.4em\relax IEEE, 2018, pp. 4802--4809.

\end{thebibliography}

% \newpage

 

% \end{IEEEbiography}



% \vfill

\end{document}